\documentclass[10pt,twocolumn,letterpaper]{article}

\usepackage{cvpr}              % To produce the CAMERA-READY version
\usepackage{times}
\usepackage{epsfig}
\usepackage{graphicx}
\usepackage{amsmath}
\usepackage{amssymb}

\usepackage{float}
\usepackage{booktabs}
\usepackage{caption}
\usepackage{subcaption}
\usepackage{multirow}
\usepackage{xcolor}

\mathchardef\mhyphen="2D
\usepackage[utf8]{inputenc}
\usepackage{comment}
\usepackage{tabularx,colortbl}
\usepackage{xparse,mathtools}
\usepackage{diagbox}
\usepackage{adjustbox,lipsum}

\usepackage{tabulary,overpic}
\newcommand{\tablestyle}[2]{\setlength{\tabcolsep}{#1}\renewcommand{\arraystretch}{#2}\centering\footnotesize}
\newlength\savewidth

\usepackage{siunitx}
\usepackage[pagebackref=true,breaklinks=true,colorlinks,bookmarks=false]{hyperref}

\newcommand{\bI}{\mathbf{I}}

\newcommand{\bx}{\mathbf{x}}
\newcommand{\by}{\mathbf{y}}

\newcommand{\btheta}{\boldsymbol{\theta}}

\newcommand{\bsigma}{\boldsymbol{\sigma}}

\makeatletter
\DeclareRobustCommand\onedot{\futurelet\@let@token\@onedot}
\def\@onedot{\ifx\@let@token.\else.\null\fi\xspace}
\def\eg{e.g\onedot} 
\def\ie{i.e\onedot}

\def\etal{et~al\onedot}
\def\iid{i.i.d\onedot}
\makeatother

\newcommand{\boldparagraph}[1]{\vspace{0.2cm}\noindent{\bf #1:}}

\definecolor{darkgreen}{rgb}{0,0.7,0}

\DeclareMathOperator{\E}{\mathbb{E}}
\DeclareMathOperator*{\minimize}{minimize}

\usepackage[capitalize]{cleveref}
\crefname{section}{Sec.}{Secs.}
\Crefname{section}{Section}{Sections}
\Crefname{table}{Table}{Tables}
\crefname{table}{Tab.}{Tabs.}

\def\cvprPaperID{2537} % *** Enter the CVPR Paper ID here
\def\confName{CVPR}
\def\confYear{2022}

\begin{document}

%%%%%%%%% TITLE - PLEASE UPDATE
\title{SelfD: Self-Learning Large-Scale Driving Policies From the Web}

\author{Jimuyang Zhang \quad Ruizhao Zhu \quad Eshed Ohn-Bar\\
	Boston University\\
	{\tt\small \{zhangjim, rzhu, eohnbar\}@bu.edu}
}

\twocolumn[{%
\renewcommand\twocolumn[1][]{#1}%
\maketitle
\thispagestyle{empty}
\begin{center}
    \centering
    \vspace{-0.6cm}
  \includegraphics[trim=6cm 6cm 12cm 8cm,clip,width=6.9in]{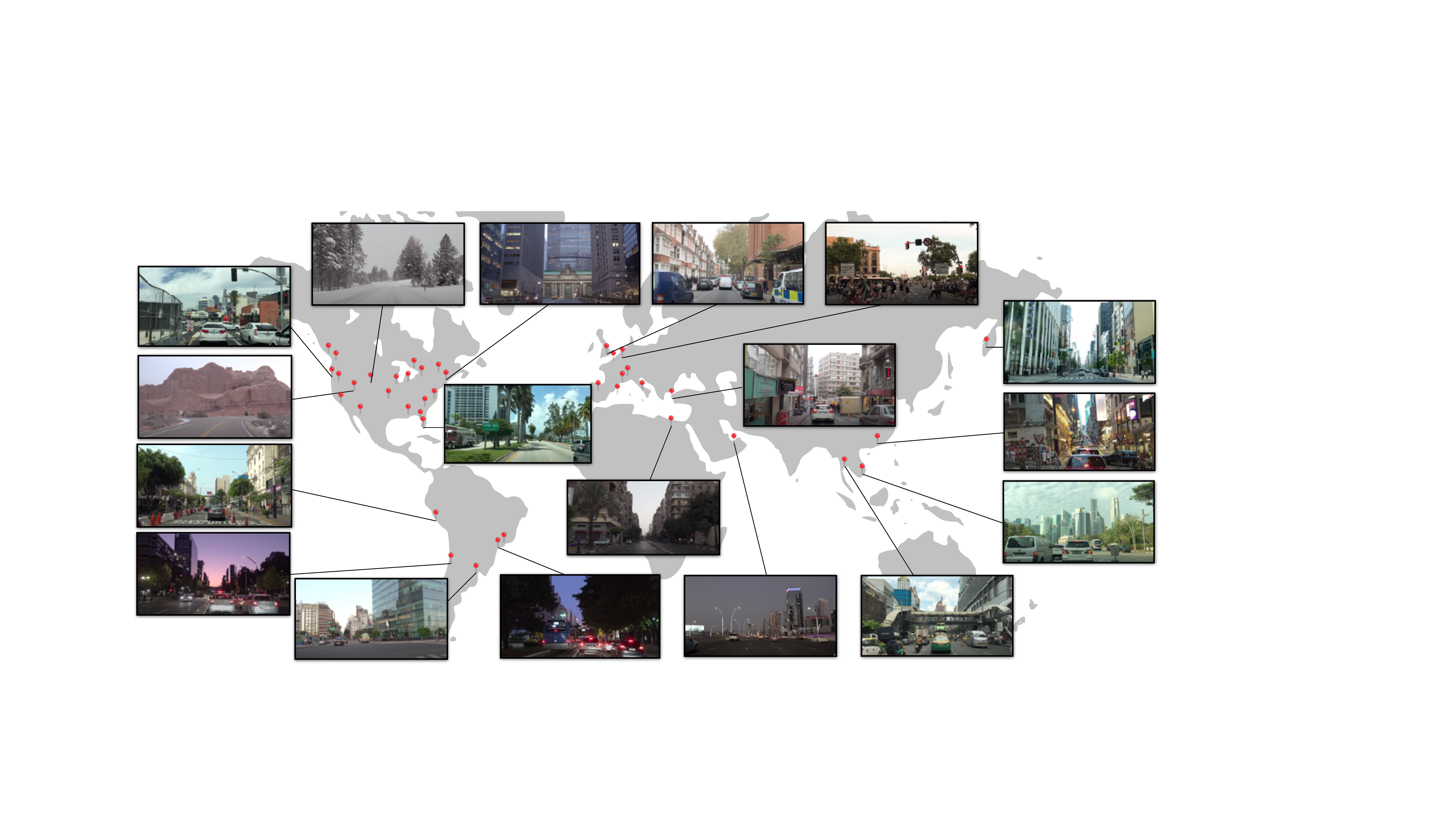}
    \vspace{-1cm}
    \captionof{figure}{\textbf{Learning From Unlabeled Data Facilitates Scalable Navigation Decision-Making.} Our goal is to develop robust, generalized, and easily deployable decision-making policies for navigation. Our key insight is to make use of the hours of freely available and highly diverse navigation data from the web in order to augment the knowledge and robustness of an initially trained navigation policy. }
    \vspace{-0.05cm}
    \label{fig:ftop}
\end{center}
}]

\begin{comment}

\end{comment}

\begin{abstract}
%%%%%%%%% ABSTRACT
Effectively utilizing the vast amounts of ego-centric navigation data that is freely available on the internet can advance generalized intelligent systems, \ie, to robustly scale across perspectives, platforms, environmental conditions, scenarios, and geographical locations. However, it is difficult to directly leverage such large amounts of unlabeled and highly diverse data for complex 3D reasoning and planning tasks. Consequently, researchers have primarily focused on its use for various auxiliary pixel- and image-level computer vision tasks that do not consider an ultimate navigational objective. 
In this work, we introduce SelfD, a framework for learning scalable driving by utilizing large amounts of online monocular images. Our key idea is to leverage iterative semi-supervised training when learning imitative agents from unlabeled data. To handle unconstrained viewpoints, scenes, and camera parameters, we train an image-based model that directly learns to plan in the Bird's Eye View (BEV) space.
Next, we use unlabeled data to augment the decision-making knowledge and robustness of an initially trained model via self-training. 
In particular, we propose a pseudo-labeling step which enables making full use of highly diverse demonstration data through ``hypothetical'' planning-based data augmentation. We employ a large dataset of publicly available YouTube videos to train SelfD and comprehensively analyze its generalization benefits across challenging navigation scenarios. Without requiring any additional data collection or annotation efforts, SelfD demonstrates consistent improvements (by up to 24\%) in driving performance evaluation on nuScenes, Argoverse, Waymo, and CARLA.
\end{abstract}

%------------------------------------------------------------------------

\vspace{-0.5cm}
\section{Introduction}
How can we learn generalized models for robust vision-based navigation in complex and dynamic settings? While humans can \textit{effortlessly transfer general navigation knowledge} across settings and platforms~\cite{kaufmann2019beauty,mishkin2019benchmarking,pan2020imitation,osa2018algorithmic}, current real-world development of navigation agents generally deploys within a fixed pre-assumed setting (\eg, geographical location, use-case) and carefully calibrated sensor configurations. Consequently, each autonomy use-case generally requires its own prohibitive data collection and platform-specific annotation efforts~\cite{bansal2018chauffeurnet,coors2019nova,pomerleau1989alvinn,bojarski2016end,xworld}. Due to such development bottlenecks, brittle navigation models trained in-house by various developers (\eg, Tesla’s Autopilot\cite{tesla}, Waymo’s Driver~\cite{bansal2018chauffeurnet}, Amazon’s Astro~\cite{Amazon}, Fedex’s Roxo\cite{Fedex}, etc.) are easily confounded by the immense complexity of the real-world navigation task, \eg, rare scenarios, novel social settings, geographical locations, and camera mount perturbations. Yet, every minute, a vast amount of highly diverse and freely available ego-centric navigation data containing such scenarios is uploaded to the web. In this paper, we work towards effectively utilizing such freely available demonstration data to improve the efficiency, safety, and scalability of generalized real-world navigation agents. 

There are two key challenges to employing the large amounts of unconstrained and unlabeled online data for training robust vision-based navigation policies. First, while online images may be collected in various layouts and camera settings, existing monocular-based prediction and decision-making methods tend to rely on restrictive assumptions of planar scenes~\cite{wang2019monocular,dickmanns1988dynamic} and known camera parameters~\cite{philion2020lift,wang2017deepvo,yang2020d3vo,gordon2019depth,li2019learning,ranftl2016dense,lbc}. 
Towards a dataset and platform-agnostic navigation agent, our proposed architecture does not explicitly rely on such assumptions. Second, due to safety-critical requirements, methods for learning decision-making in complex navigation settings generally also assume access to highly curated benchmarks with clean annotations~\cite{chang2019argoverse,Geiger2012CVPR,cordts2016cityscapes,houston2020one ,caesar2020nuscenes,sun2020scalability,bansal2018chauffeurnet,wu2019imitation,Codevilla2018ICRA,lbc}. Consequently, such methods must be revisited when learning from unlabeled and diverse internet videos, \eg, with various quality demonstrations~\cite{wu2019imitation}. To effectively utilize such data, we propose to leverage recent advances in iterative semi-supervised learning~\cite{lee2013pseudo,rizve2021defense,caine2021pseudo}. Yet, as such techniques are studied within pixel and image-level tasks~\cite{chen2015webly,li2019learning,goyal2019scaling}, their utility for learning complex and safety-critical 3D reasoning and planning tasks is not well-understood.

\boldparagraph{Contributions} Motivated by how humans are able to effortlessly learn and adapt various skills using online videos, our SelfD approach introduces three main contributions. \textit{First}, to facilitate learning from unconstrained imagery, we develop a model for mapping monocular observations directly to a Bird's Eye View (BEV) planning space (\ie, without requiring camera calibration). \textit{Second}, we introduce a novel semi-supervised learning approach incorporating self-training with ``hypothetical'' data augmentation. 
We demonstrate the proposed sampling step to be essential for making use of highly diverse demonstration data.
\textit{Third}, we perform a set of novel cross-dataset experiments to analyze the ability of an initially trained decision-making model to self-improve its generalization capabilities using minimal assumptions regarding the underlying data. We evaluate across various datasets with complex navigation and harsh conditions to demonstrate state-of-the-art model generalization performance.
\section{Related Work}
\label{subsec:relatedwork}

\boldparagraph{Observational Imitation Learning} Our key idea is to leverage the scale and diversity of easily available online ego-centric navigation data to learn a robust conditional imitation learning policy~\cite{Codevilla2018ICRA,lbc}. 
% decision-making 
While learning from labeled demonstrations can significantly simplify the challenging vision-based policy learning task~\cite{zhang2021learning,Zhoueaaw6661,bojarski2016end,osa2018algorithmic,li2018oil,zhao2019lates,Bain96aframework,pomerleau1989alvinn,muller2006off,Chen2015ICCVa,muller2018driving,liang2018cirl,gupta2017cognitive,prakash2021multi,chen2021learning,hawke2020urban}, observed images in our settings are not labeled with corresponding actions of a demonstrator. We therefore work to generalize current conditional imitation learning (CIL) approaches~\cite{lbc,Codevilla2018ICRA,Codevilla2019ICCV} to learn, from unlabeled image observations, an agent that can navigate in complex urban scenarios. 
To address this challenging observational learning task, prior work has recently explored introducing various restrictive assumptions, including access to a hand-designed reward function~\cite{chang2020semanticyt}, an interactive environment for on-policy data collection~\cite{torabi2018behavioral}, or demonstrator optimality~\cite{torabi2018behavioral,torabi2019recent}. Moreover, due to unconstrained and dynamic scenes in our settings, learning an inverse model~\cite{Sun-2019-118906,torabi2018behavioral,torabi2019recent,chang2020semanticyt} (\ie, the visual odometry task of recovering actions from observation sequences) becomes challenging and may result in noisy trajectories. To facilitate scalable training from diverse data sources, our proposed semi-supervised learning approach does not leverage such assumptions. Nonetheless, our resulting model can also be used to bootstrap other methods for policy training, \eg, model-based or model-free reinforcement learning approaches~\cite{ohn2020learning,liang2018cirl,toromanoff2020end,chen2021learning}. 

\boldparagraph{Semi-Supervised Learning for Navigation}
CIL generally involves learning from known actions of human experts~\cite{Codevilla2018ICRA,Codevilla2019ICCV,prakash2021multi}, which is not applicable to our settings. Yet, the recent work of Chen~\etal~\cite{lbc} utilizes a multi-stage training step, where a privileged (\ie, `teacher') CIL agent is employed to provide supervision to a non-privileged (\ie, `student') visuomotor CIL agent. As the privileged agent is given access to extensive ground truth information about the world in training and testing, it produces highly plausible and clean trajectories.
In contrast, SelfD leverages \textit{the same visuomotor architecture} as teacher and student. We also train in inherently noisy settings, as teacher inference is performed on diverse out-of-distribution image data and not on the original training dataset. Consequently, our approach goes beyond prior policy distillation approaches to handle scenarios where supervision by the teacher model may be potentially noisy and unsafe. 
We also note the relationship between such distillation and semi-supervised training via pseudo-labeling ~\cite{caine2021pseudo,lee2013pseudo,rizve2021defense,saito2017asymmetric,lin2021streaming}. However, as far as we are aware, we are the first to develop a pseudo-labeling based self-training method for learning safe driving policies from complex scenes with diverse navigation data, camera perspectives, geographical locations, and weathers. In contrast, prior work~\cite{lee2013pseudo,rizve2021defense} emphasizing semi-supervised learning for image- and object-level recognition tasks~\cite{yang2021st3d,caine2021pseudo} has limited utility for complex decision-making tasks as further discussed in Sec.~\ref{sec:experiments}.

\boldparagraph{Self-Supervised Visual Learning}
Our approach aims to directly leverage experience from large amounts of unlabeled video data to learn complex 3D navigation. An alternative approach to our proposed semi-supervised framework involves learning visual representations from the large unlabeled data~\cite{goyal2019scaling,sohn2020fixmatch,he2020momentum,patrick2021compositions,godard2019digging,chen2019self,liu2019selflow,liu2020flow2stereo}. These generic representations can then be transferred to the policy learning task~\cite{goyal2019scaling}. However, visual representation learning from unlabeled data often rely on various auxiliary image-level data augmentation strategies~\cite{asano2020labelling,watson2021temporal,noroozi2016unsupervised,zhang2016colorful}, \eg, Jigsaw or Colorization~\cite{goyal2019scaling} tasks, that are indirect to the ultimate navigation task. Thus, such methods are limited in utility for elaborate spatial navigation tasks in dynamic settings. Indeed, the utility of image augmentation-based methods beyond navigation in simple static environments (\eg,~\cite{goyal2019scaling}) has yet to be demonstrated, \ie, for autonomous driving. In contrast, SelfD's learning and augmentation strategies directly optimize for BEV (map) planning. 
Moreover, our approach is orthogonal to the aforementioned studies as in-direct visual tasks can be leveraged as further pre-training prior to the proposed self-supervised pseudo-labeling step.

%\vspace{-0.1in}
\section{Method}
\label{sec:method}
Our goal is to facilitate training driving policies at scale. To efficiently make use of the broad and diverse experience found in large amounts of unlabeled videos, we follow three main steps. We first propose to use a monocular image-based behavior cloning planner that reasons directly in the BEV (Sec.~\ref{subsec:BEV}). Our proposed planner can therefore better generalize across arbitrary perspectives.
Next, we introduce a data augmentation step for obtaining multiple plausible pseudo-labels when self-training over unlabeled internet data (Sec.~\ref{subsec:pl}). Finally, we re-train the model over the larger dataset to learn a more robust and generalized vision-based navigation policy (Sec.~\ref{subsec:pl2}).

\subsection{Problem Setting} 
\label{subsec:define}
We consider the task of learning from observations $\bx = (\bI, v, c) \in \mathcal{X}$ of a single front camera image $\bI\in\mathbb{R}^{W\times H\times 3}$, ego-vehicle speed $v\in\mathbb{R}$, and a categorical navigational command $c\in\mathbb{N}$ (\eg, turn left, turn right, and forward~\cite{Codevilla2018ICRA}). Our agent learns to map such observations to a navigational decision. In general, the decision may either be a low-level vehicle control action~\cite{Codevilla2019ICCV} (\eg, steering, throttle) or a desired future trajectory relative to the ego-vehicle, \ie, set of $K$ waypoints $\by \in \mathcal{Y}$~\cite{muller2018driving,lbc} in the BEV (map) space. 
In the latter case, future waypoints may be paired with a hand-specified or learned motion controller to produce the low-level action~\cite{lbc,muller2018driving}. 
In this work, we focus on the latter representation due to its interpretability and generalizability~\cite{muller2018driving}. Thus, our goal is to obtain a waypoint prediction function $f_{\btheta}\colon\mathcal{X} \to \mathcal{Y}$, with learnable parameters $\btheta~\in~\mathbb{R}^d$, \ie, the decision-making policy. Without loss of generality, we will slightly augment the output space of this function in Sec.~\ref{subsec:BEV}. Next, we build on recent advances in driving policy learning, in particular CIL~\cite{Codevilla2019ICCV,Codevilla2018ICRA,lbc}, to develop a general approach for learning to drive from unlabeled data.

\boldparagraph{Conditional Imitation Learning} 
Navigational demonstrations can be collected by logging the sensor data of a manually operated system, as often done by researchers and developers. Such demonstrations can then be utilized for training decision-making policies through various learning techniques~\cite{choudhury2018data,abbeel2004apprenticeship,Codevilla2018ICRA,pomerleau1989alvinn,bansal2018chauffeurnet,osa2018algorithmic,ross2011reduction}. In its most straightforward implementation, training the mapping function $f_{\btheta}$ can be simplified to supervised learning via an \iid data assumption~\cite{pomerleau1989alvinn,behl2020label}.
More robust policies can be trained by going beyond such schemes, \eg, through online and interactive learning strategies~\cite{osa2018algorithmic,ross2011reduction,Prakash2020CVPR}. In our work, we leverage off-line demonstrations and do not assume access to an interactive environment. Nonetheless, our proposed data augmentation techniques in Sec.~\ref{subsec:pl} can significantly improve policy robustness, as will be demonstrated in Sec.~\ref{sec:experiments}.

Given a collected dataset of observations and corresponding actions, $\mathcal{D} = \{(\bx_i,\by_i) \}_{i=1}^{N}$,  supervised training can be achieved by optimizing an imitation objective
\begin{equation}
\label{eqn:lossil}
\minimize_{\btheta} \E_{(\bx,\by) \sim \mathcal{D}} \left[ \mathcal{L} (\by, f_{\btheta}(\bx)) \right] 
\end{equation}
with $\mathcal{L}$ being a suitable loss function, \eg, $L_1$ regression distance to the waypoint targets. Despite widespread use in academia~\cite{hu2021safe,Codevilla2019ICCV,xu2017end} and industry~\cite{bansal2018chauffeurnet,bojarski2016end}, the standard CIL formulation significantly limits scalability. Specifically, approaches for imitation learning generally require having direct access to optimal target action labels. Consequently, they cannot be applied towards learning from the vast and diverse unlabeled data that is uploaded online. 

\boldparagraph{Conditional Imitation Learning from Observations}
To make full use of unlabeled data in demonstrations containing diverse navigational experience, we develop a framework for Conditional Imitation Learning from Observations (CILfO), \ie, assuming only data of the form $\mathcal{U} = \{\bI_i \}_{i=1}^{M}$. Within this more generalized yet difficult learning task, a key challenge lies in recovering suitable label targets $\hat{\by}$, navigational command $\hat{c}$ and speed $\hat{v}$ to construct a dataset 
\begin{equation}
    \hat{\mathcal{D}} = \{(\bI_i, \hat{v}_i, \hat{c}_i), \hat{\by}_i \}_{i=1}^{M}
\end{equation}
over which Eqn.~\ref{eqn:lossil} can then be used to train a policy. 

\boldparagraph{Initial Data Assumption} In order to address the challenging CILfO learning task, our key idea is to leverage a small labeled dataset to learn an initial policy mapping using human expert demonstrations. We then sample from this trained function to obtain pseudo-labels~\cite{lee2013pseudo} over the unlabeled data.
This assumption is reasonable considering that there are several publicly available driving datasets with included action labels~\cite{wilson2021argoverse,caesar2020nuscenes,houston2020one,Geiger2012CVPR}. Alternatively, a preliminary dataset with a novel platform setup or use-case may be initially collected. 

\boldparagraph{Sequential Data Assumption} We note that although sequential video data is used in our ablation, our generic and scalable CILfO formulation does not assume access to temporal data. The reason for this is three-fold. First, while sequential observations may be potentially beneficial for disambiguation of various decision-making factors~\cite{de2019causal}, generalization results for learning temporal sensorimotor driving policies have been inconclusive~\cite{de2019causal,muller2006off,wang2019monocular}. Second, while sequential data is assumed in most prior observational imitation learning settings~\cite{chang2020semanticyt,torabi2018behavioral} recovering the underlying demonstration actions (\eg, using monocular visual odometry~\cite{wang2017deepvo}) in arbitrary scenes is challenging. Due to the difficulty and noise in sequential action recovery, our proposed single-frame formulation in Sec.~\ref{subsec:drivenet} is shown to significantly outperform such baselines in Sec.~\ref{sec:experiments}. Third, we do not make any assumptions regarding the optimality of the demonstrations in the unlabeled dataset $\mathcal{U}$~\cite{murali2015learning,torabi2019recent}.
Consequently, our generalized method is suitable for scalable learning from videos with complex scenes, environmental conditions, arbitrary viewpoints, and diverse demonstration quality. As our learning task lies well beyond the capabilities of existing CIL and observational methods, we next develop a generalized training method for leveraging unconstrained and unlabeled demonstration data.

\subsection{SelfD Overview} 
\label{subsec:drivenet}

Through a semi-supervised policy training process, our proposed SelfD navigation policy model can be learned in three summarized steps:

\begin{enumerate}
	\setlength\itemsep{0em}
	\item Use a small, labeled domain-specific dataset $\mathcal{D}$ to learn an initial observations-to-BEV policy $f_{\btheta}$ via imitation.
	\item Obtain a large pseudo-labeled dataset $\hat{\mathcal{D}}$ by leveraging sampling from $f_{\btheta}$.
	\item Pre-train a generalized policy $f_{\btheta}$ on $\hat{\mathcal{D}}$ and fine-tune on the clean labels of $\mathcal{D}$. 
\end{enumerate}

Note that we re-use the parameter symbol ${\btheta}$ throughout the steps to simplify notation.
Our iterative semi-supervised training enables effectively augmenting the knowledge and robustness of an initially trained policy. As described next, our proposed initial step facilitates learning a platform and perspective-agnostic policy during subsequent training steps by directly reasoning in a BEV planning space. 

\subsection{BEV Plan Network}
\label{subsec:BEV}
In this section, we propose a suitable output representation to account for arbitrary cameras, viewpoints and scene layouts. Current monocular planners generally predict waypoints in the image plane to align with an input image~\cite{muller2018driving,lbc}. The waypoints are then transformed to a BEV plan using carefully calibrated camera intrinsic and extrinsic (\eg, rotation, height) parameters~\cite{lbc}. Thus, policy models are often trained and evaluated within a fixed pre-assumed setup. In contrast, SelfD predicts a future plan parameterized by waypoints in the BEV plan space directly. Based on our experiments in Sec.~\ref{sec:experiments}, we demonstrate this choice to be crucial for real-world planning across settings. The predicted generalized BEV waypoints can be paired with a low-level controller, \eg, a PID controller~\cite{lbc,muller2018driving}.
Due to the difficulty in learning a monocular-to-BEV plan mapping, we follow recent work in confidence-aware learning~\cite{kendall2018multi} to train an augmented model $f_{\btheta}\colon\mathcal{X} \to \mathcal{Y} \times \mathcal{R}$ with quality estimates $\bsigma \in \mathcal{R}$.
Our training loss function in Eqn.~\ref{eqn:lossil} is defined as
\begin{equation}
    \mathcal{L} = \mathcal{L}_{\text{plan}} +\lambda\mathcal{L}_{\text{quality}}
\label{eq:lambda}
\end{equation}
where $\mathcal{L}_{\text{plan}}$ is the $L_1$ distance between ground-truth and predicted waypoints, $\mathcal{L}_{\text{quality}}$ is a binary cross-entropy loss~\cite{yang2021st3d,kendall2018multi}, and the $\lambda$ hyper-parameter balances the tasks.
\subsection{``What If'' Pseudo-Labeling of Unlabeled Data}
\label{subsec:pl}
\begin{figure}[!t] %!t
           \centering
       \includegraphics[trim=0cm 2.2cm 2.4cm 0cm, width=3.3in]{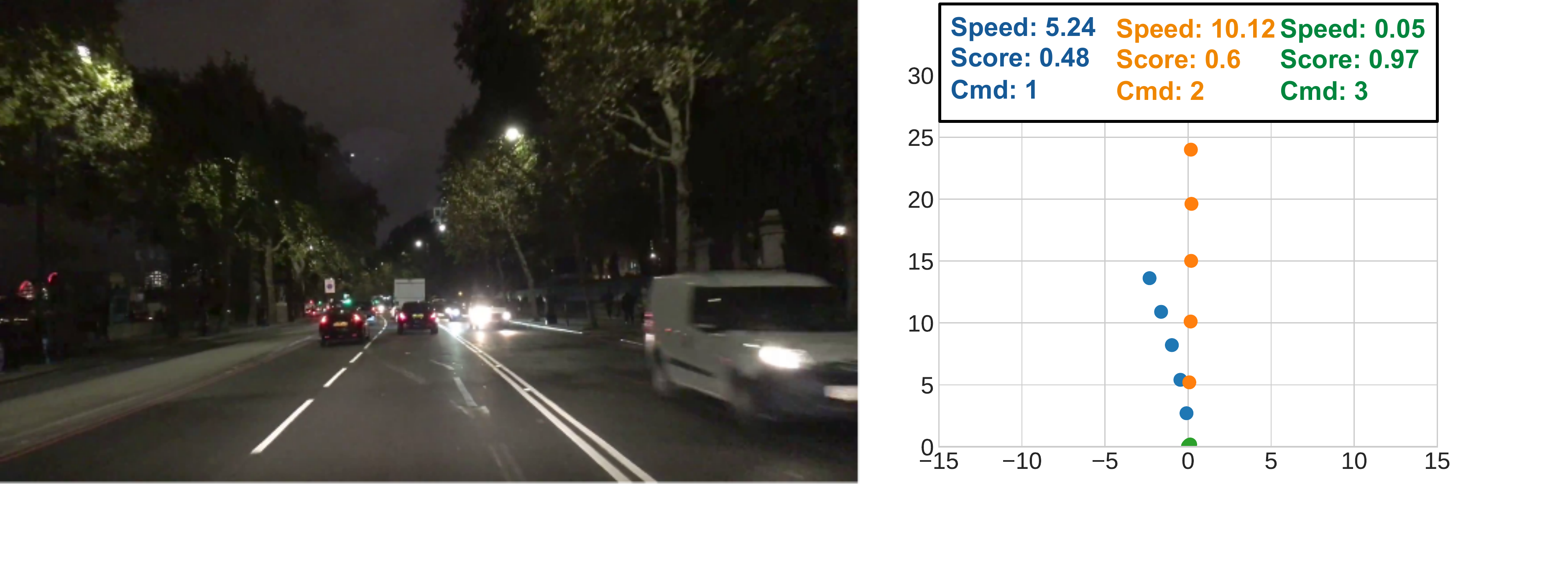} \\
       \includegraphics[trim=0cm 2.2cm 2.4cm 0cm, width=3.3in]{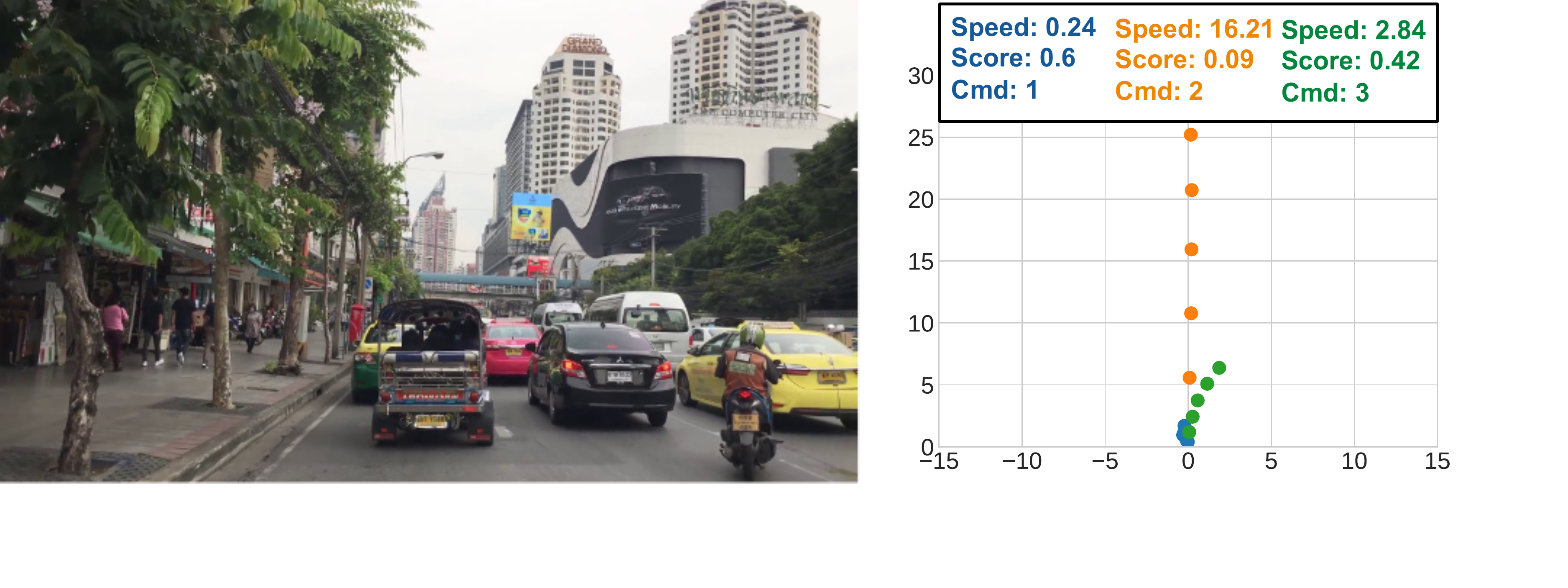}
      \caption{\textbf{``What If'' Hypothetical Pseudo-Labeling.} We generate multiple plausible future trajectories (\textbf{units are meters} in BEV) as pseudo-labels for each unlabeled frame by sampling the conditional planner from Sec.~\ref{subsec:BEV}. We depict two scenarios from our dataset with inference shown for various inputs. Speed is in meters per second and conditional commands are either left (1), forward (2), or right (3).  }
   \label{fig:plfig}
  \end{figure}

Given a set of unlabeled images $\mathcal{U}$, we sample from the trained conditional policy $f_{\btheta}$ in a semi-supervised training process. While the speed and command inputs to $f_{\btheta}$ can be recovered through visual odometry techniques~\cite{wang2017deepvo}, these result in highly noisy trajectories in our online video settings (discussed in Sec.~\ref{sec:experiments}). As the demonstrations in our data may be unsafe or difficult to recover, we propose to leverage a single-frame pseudo-labeling mechanism. Our key insight is to employ the conditional model $f_{\btheta}$ to generate multiple hypothetical future trajectories in a process referred to as ``what if'' augmentation. 
Beyond resolving the missing speed and command inputs, our proposed augmentation provides additional supervision, \ie, a conditional agent that better reasons on what it \textit{might need to do}, for instance, if it had to turn left instead of right at an intersection (Fig.~\ref{fig:plfig}). In contrast to related work in policy learning and distillation~\cite{lbc}, sampling from the teacher agent is more challenging as the agent is not exposed to extensive 3D perception knowledge about the world and is being evaluated outside of its training settings. 

We repeatedly sample $\hat{v}$ and $\hat{c}$ uniformly and rely on the conditional model to provide pseudo-labels $(\hat{\by}, \hat{\bsigma}) = f_{\btheta}(\bI,\hat{v},\hat{c})$ for additional supervision across all conditional branches and speed observations. 
In this manner, querying the ``teacher'' model $f_{\btheta}$ enables us to generate various scenarios beyond the original demonstration. In particular, as discussed in Sec.~\ref{sec:experiments}, we find self-training strategies to provide limited generalization gains without this ``what if'' data augmentation step. This augmentation strategy enables our single-frame pseudo-labeling approach to significantly outperform approaches that are more elaborate to train at scale, as they may involve additional modules relying on approximating $\hat{\by}$, $\hat{c}$, and $\hat{v}$ from video. Finally, to avoid incorporating potentially noisy trajectories, the corresponding quality estimates $\hat{\bsigma}$ can be used to process and filter examples in the pseudo-labeled dataset $\hat{\mathcal{D}}$.

\subsection{Model Pre-Training and Fine-Tuning} 
\label{subsec:pl2}
As a final training step, we re-train the waypoint network $f_{\btheta}$ from scratch over the large and diverse dataset $\hat{\mathcal{D}}$.
The pre-trained policy can then be further fine-tuned over the original dataset $\mathcal{D}$, thus leveraging the additional knowledge gained from $\hat{\mathcal{D}}$ to improve its performance. We note that we employ separate training over the two datasets $\hat{\mathcal{D}}$ and $\mathcal{D}$ and rely on knowledge transfer through learned representations as it reduces the need for any careful hyperparameter tuning beyond the overall learning rate. For instance, Caine~\etal~\cite{caine2021pseudo} empirically demonstrate the importance of delicately optimizing the ratio of labeled to pseudo-labeled data when mixing the datasets for a 3D object detection task, while also showing it to vary among object categories, \eg, pedestrians vs. vehicles. Thus, through a pre-training mechanism, we avoid the need to carefully mix the cleanly labeled and pseudo-labeled datasets~\cite{sohn2020fixmatch,caine2021pseudo,yang2021st3d}.

\subsection{Implementation Details} 
\label{subsec:implement}
We implement our BEV waypoint prediction network $f_{\btheta}$ leveraging a state-of-the-art conditional sensorimotor agent~\cite{lbc}. However, as discussed in Sec.~\ref{subsec:drivenet}, we do not assume a fixed known BEV perspective transform. Thus, we remove the fixed perspective transformation layer which restricts scalability and replace it with a per-branch BEV prediction module (see supplementary for additional implementation and architecture details). During training we use a learning rate of \num{1e-3} and resize images to $400\times225$. 

\section{Experiments}
\label{sec:experiments}

\subsection{Experimental Setup}
\label{sec:expsetup}
To obtain large amounts of driving data, we downloaded 100 hours of front view driving videos from popular driving channels on YouTube (across cities, weathers, and times of day, as shown in Fig.~\ref{fig:ftop}). In our real-world evaluation, we use the nuScenes~\cite{caesar2020nuscenes}, Waymo~\cite{sun2020scalability} and Argoverse~\cite{chang2019argoverse} datasets. While nuScenes is a highly instrumented and annotated dataset, it is primarily used for perception tasks without an official benchmark for future waypoint planning. The recent work by Hu~\etal~\cite{hu2020you} uses a random split, yet this is not suitable for studying generalization (the method also employs LiDAR input whereas we do not). We create a geography-based split into nuScenes Boston (nS-Boston) and nuScenes Singapore (nS-Singapore) which is challenging due to the significant domain shift. 
To further evaluate generalization, we also utilize the Waymo and Argoverse datasets which were collected across 8 different cities. Specifically, we construct a future waypoint prediction benchmark from the Waymo perception testing set and the Argoverse 3D tracking training set. Note that in our evaluation we do not assume any access to the target domain's pseudo-labels, as often done in related research. Overall, our split results in $13K$, $11K$, $11K$, and $26K$ samples from nS-Boston, nS-Singapore, Argoverse and Waymo, respectively. 

The open-loop real-world evaluation compares predictions to an expert driver in complex interactions (\ie, diverse maneuvers, yielding, merging, irregular intersections). To analyze the impact of the proposed approach during closed-loop driving, we further perform interactive on-policy evaluation in CARLA~\cite{Dosovitskiy2017CORL,Codevilla2019ICCV}. We replicate our training settings in simulation and predict a final low-level action by employing a PID controller~\cite{lbc}. In particular, we train on labeled data from Town 1 and evaluate on Town 2 in regular traffic (following NoCrash~\cite{Codevilla2019ICCV}). To understand the impact of pseudo-labeling on the close-loop driving metrics we keep a portion of the training town data for pseudo-labeling. The evaluation is conducted over 25 predefined routes in Town 2 under four different weathers, one of which is not seen in annotated data. Additional details regarding the experimental setup can be found in the supplementary.

\boldparagraph{Evaluation Metrics}
We follow standard open-loop evaluation and use Average and Final $L_2$ Displacement Error (ADE and FDE, respectively) over future waypoints in the BEV~\cite{li2020end}. We also compute a collision rate~\cite{hu2021safe} which measures collisions along the predicted waypoints with other vehicles (we are only able to compute this metric on nuScenes and Argoverse due to the provided annotations for the surrounding vehicles). In our CARLA experiments we also report agent Success Rate (SR), Route Completion (RC), and Collision (Coll.) frequency per $10$km. 

\subsection{Results} 
\label{subsec:results}

\boldparagraph{Model Architecture}
We first analyze the proposed BEV planning model architecture on the nuScenes cross-town split. The architecture is then fixed for subsequent experiments. As shown in Table~\ref{tab:architecture}, the state-of-the-art CIL model~\cite{lbc} achieves a 1.86 ADE on the cross-town training and testing split. We emphasize that the CIL baseline~\cite{lbc} predicts the waypoints in 2D images and transform the predicted waypoints into BEV through a perspective transformation. However, such transformation may not be accurate in the real-world. We can therefore see the benefits of the BEV planner, even before pseudo-labeling techniques have been applied. Our proposed BEV planner achieves a 1.14 ADE (a nearly $40\%$ reduction in error compared to CIL). Given prior work on nuScenes primarily relies on non-conditional approaches based on LiDAR~\cite{hu2021safe}, we also ablate over different inputs to the model. Intuitively, we find the speed measurements to be critical to BEV waypoint prediction. Our proposed BEV planner with both speed and conditional command input gives the best ADE performance on nS-Singapore. We now continue to analyze the benefits of a semi-supervised learning step to improve policy performance.

\boldparagraph{Pseudo-Labeling Approaches}
We consider various approaches for leveraging the unlabeled YouTube data (we select 10 driving hours for the ablation). To emphasize generalization, we do not pseudo-label the unseen test datasets or incorporate their unlabeled samples into the self-training in any way. We report waypoint prediction performance before and after the final fine-tuning step on nS-Boston across all test sets. 
As shown in Table~\ref{tab:main}, pre-training over the pseudo-labeled YouTube data and fine-tuning on the clean dataset consistently improves ADE, FDE, and collision metrics across evaluation settings. We see how leveraging a state-of-the-art Visual Odometry (VO) model~\cite{wang2017deepvo} as a teacher for SelfD instead of the proposed pseudo-labeling mechanism results in reduced performance. As recovering the camera motion in dynamic scenes with unknown camera parameters is difficult, leveraging this model results in noisy and unsafe trajectories. While we find limited utility in leveraging temporal data, this may be alleviated by improved visual odometry methods in the future. As such methods become more robust to arbitrary settings, they may be integrated to complement our method, \eg, to leverage both rollouts as well as the proposed data augmentation. We also find our pre-training pipeline to improve over mixing-based approaches~\cite{sohn2020fixmatch}.

 \begin{table}[!t]
 \caption{\textbf{Model Architecture.} The CIL baseline~\cite{lbc} predicts waypoints in the image and relies on perspective transform with known camera calibration, while our proposed monocular-to-BEV planner does not have access to camera intrinsic or extrinsic parameters. All models are trained on nuScenes Boston and evaluated on nuScenes Singapore. }
 \label{tab:architecture}
     \centering
     \noindent\adjustbox{max width=\columnwidth}{
     \begin{tabular}{l| c c |c }
     \toprule
      Models & Command & Speed & ADE (meters) \\
      \midrule
     CIL Baseline~\cite{lbc} &\checkmark & \checkmark&  1.86 \\ 
     \midrule
     \multirow{4}{*}{\begin{tabular}{l}BEV Planner
                \end{tabular}} 
                &  &  & 4.77 \\
                & \checkmark  & & 4.64 \\
                & &\checkmark   &  1.33 \\
                & \checkmark & \checkmark & \textbf{1.14} \\
     \bottomrule
     \end{tabular}}
 \end{table}

\begin{table*}[t]
\tablestyle{8pt}{1.1}
\caption{\textbf{Pseudo-Labeling Approaches.} We compare various pseudo-labeling approaches by varying the underlying sampling mechanism for the speed and conditional inputs. These may either be estimated with visual odometry~\cite{wang2017deepvo} (\ie, SelfD without ``what if'' augmentation) or sampled from a prior distribution (\ie, uniformly or with a prior). We find uniform sampling to work best as a form of data augmentation. We also compare to a baseline using a state-of-the-art visual odometry method for pseudo-labels. Note that DVO-based approaches utilize sequential video data while the rest are single-frame. ``Coll.'' refers to collision rate and ``nS'' to nuScenes (see Sec.~\ref{sec:expsetup}).
}
\label{tab:main}
\begin{tabular}{l l|c c c | c c c |c c c}
\toprule
\multicolumn{2}{c|}{\multirow{2}{*}{Method}} & \multicolumn{3}{c|}{nS-Singapore} & \multicolumn{3}{c|}{Argoverse} & \multicolumn{2}{c}{Waymo} \\
& & ADE & FDE & Coll. (\%) & ADE & FDE & Coll. (\%)  & ADE & FDE  \\
     \midrule
\multicolumn{2}{c|}{BEV Planner (nS-Boston)}  & 1.14 & 2.16 & 9.50 &1.07 & 2.16 &16.50 & 2.17 & 3.87  \\
  \midrule
 \multirow{4}{*}{\begin{tabular}{l}Pre-Trained \\ (YouTube)
                \end{tabular}}
&SelfD (VO) & 4.63 & 8.07 & 12.91 & 4.35 & 7.54 & 31.87 & 5.16  & 9.02   \\
&SelfD (w/o What If)   & 1.32 & 2.45 & 10.02& 1.27 & 2.46 & \textbf{13.92} & 2.43 & 4.28 \\
&SelfD (What If-Prior)  & 1.19 & 2.40 & 10.01 & 1.12 & 2.34 & 16.51 & 2.10 & 3.94   \\
&SelfD (What If-Uniform)  & 1.19 & 2.28 & 10.10 & 1.13 & 2.28 & 15.70 & 2.24 & 4.01  \\
 \midrule
\multirow{5}{*}{\begin{tabular}{l}Fine-Tuned\\(YouTube \& nS-Boston)\end{tabular}}
&{FixMatch~\cite{sohn2020fixmatch}}  & 1.18 & 2.22 & 12.02 &1.10 &2.24 & 15.20 & 2.18 & 3.88  \\
&SelfD (VO)  & 1.09 & 2.09 & 9.33 & 1.03 & 2.11 & 17.52 & 2.01  & 3.62  \\
&SelfD (w/o What If) & 1.18 & 2.23 & 9.34 & 1.09 & 2.18 & 16.26 & 2.03 & 3.62 \\
&SelfD (What If-Prior)  & 1.09 & 2.10 & 9.47 & 1.00 & 2.06 & 15.71 & 2.10& 3.73  \\
&SelfD (What If-Uniform)  & \textbf{1.00} & \textbf{1.93} & \textbf{9.30} & \textbf{0.99} & \textbf{2.05} & 14.20 & \textbf{1.65} & \textbf{3.02} \\

\bottomrule

\end{tabular}
\end{table*} 

\begin{table*}[!t]
\tablestyle{4pt}{1.1}
\caption{\textbf{Generalization to Harsh Conditions.} We analyze the impact of SelfD on generalization from nS-Boston to specific environmental conditions. We compute the performance over domain subsets within each dataset. While insightful for the larger nuScenes and Waymo splits, we note that this splits leads to a small dataset on Argoverse with limited insights. 
}
\label{tab:harsh}
\begin{tabular}{l|c c |c c |c c|c c|c c}
\toprule
\multirow{2}{*}{Method }& \multicolumn{2}{c|}{nS-Singapore Night}  & \multicolumn{2}{c|}{Argoverse Rainy} & \multicolumn{2}{c|}{Argoverse Night} & \multicolumn{2}{c|}{Waymo Rainy} & \multicolumn{2}{c}{Waymo Night}  \\
 & ADE & FDE  &  ADE & FDE   &  ADE & FDE &  ADE & FDE   &  ADE & FDE  \\
\midrule
    BEV Planner (nS-Boston) & 1.67 & 3.05  & 0.81 & \textbf{1.71} & \textbf{0.78} & \textbf{1.61} & 0.70 & \textbf{1.28} & 1.81 & 3.25  \\
  SelfD (Pre-Trained) & 1.73 & 3.17 & 0.84 & 1.85  & 0.94 & 1.94 & 0.77 & 1.42  & 1.96 & 3.55  \\
 SelfD (Fine-Tuned) & \textbf{1.28} & \textbf{2.42} & \textbf{0.79}  & \textbf{1.71}  & 0.91  & 1.80  & \textbf{0.67} & 1.32  & \textbf{1.53} & \textbf{2.80}  \\
\bottomrule
\end{tabular}
\end{table*}

\begin{table}[t]
\tablestyle{8pt}{1.1}
\caption{\textbf{Impact of Initial Training Dataset.} We analyze the impact of different datasets for training the initial BEV planner model.
}
\label{tab:diff_init}
\begin{tabular}{l l|c c }
\toprule
\multirow{2}{*}{Initial Dataset} & {\multirow{2}{*}{Method }} & \multicolumn{2}{c}{nS-Singapore}\\
& & ADE & FDE  \\
 \midrule

 \multirow{3}{*}{\begin{tabular}{l}Argoverse
                \end{tabular}}
&BEV Planner        & 1.42 & 2.58    \\
&SelfD (w/o FT)  & 1.38 & 2.52 \\
&SelfD (w/ FT)   & \bf{1.18} & \bf{2.23}  \\

 \midrule
 \multirow{3}{*}{\begin{tabular}{l}Waymo
                \end{tabular}}
&BEV Planner         & 1.08 & 2.03 \\
&SelfD (w/o FT)   & 1.13 & 2.14 \\
&SelfD (w/ FT)    & \bf{0.91} & \bf{1.81}  \\
\bottomrule
\end{tabular}
\end{table} 
 \begin{table}[!t]
    \tablestyle{3pt}{1.1}
    \caption{\textbf{Closed-Loop Evaluation in CARLA.} We analyze the impact of the proposed self-training step on closed-loop metrics of Success Rate (SR), Route Completion (RC), and Collision frequency (Coll.) per $10$km. }
     \label{tab:carla}
     \centering
     \noindent\adjustbox{max width=1.8\columnwidth}{
     \begin{tabular}{l| c  |c | c  |c |c}
     \toprule
      Method & ADE $\downarrow$ & FDE $\downarrow$ &  SR ($\%$) $\uparrow$ & RC ($\%$) $\uparrow$ & Coll. $\downarrow$\\ 
      \midrule
     CIL Baseline~\cite{lbc}  & $0.57$	&  $1.30$ & $16$	&  $66.88$ &  $10.71$ 	\\ 
     BEV Planner & $0.66$	&  $1.32$ & $12$	&  $65.66$ &  $11.33$ 	\\ 
     \midrule
    SelfD (Town 1) & $0.56$	&  $1.30$ & $25$	&  $71.84$ &  $11.51$	\\ 
    SelfD (Town 1 \& 2)& $\bf{0.55}$	&  $\bf{1.24}$  & $\bf{26}$	&  $\bf {74.50}$ &  $\bf {5.18}$	\\ 
     \bottomrule
     \end{tabular}}
 \end{table}

\boldparagraph{What If Augmentation} Results in Table~\ref{tab:main} demonstrate how self-training without the proposed data augmentation results in limited gains. Here, we experiment with pseudo-labeling using various sampling strategies for speed measurements and conditional commands. In addition to uniform and prior-based sampling (determined empirically based on the training data), we also report findings using the visual odometry~\cite{wang2017deepvo} model estimates for the speed and conditional command. We find uniform sampling to outperform more informed prior and visual odometry-based sampling due to the extensive speed profile and multi-branch supervision it provides. We can see how a model learned through self-training but without such powerful augmentation (\ie, SelfD without What if in Table~\ref{tab:main}) results in worse performance compared to the baseline BEV planning model, \eg, 1.18 vs. 1.14 ADE and 2.23 vs. 2.16 FDE. We also find the augmentation to improve safety of the predicted trajectories, as shown by reduced collision rates. This is affirmed in our closed-loop evaluation shown in Table~\ref{tab:carla}.

\boldparagraph{Generalization to Harsh Settings} We stress test SelfD in Table~\ref{tab:harsh} under the most challenging subsets across the datasets, as selected by video episodes collected in certain environmental conditions. Interestingly, we find SelfD to efficiently reduce error on nighttime conditions, which is only seen in YouTube videos. However, as nS-Boston already contains some scenes with cloudy and light rain settings, the improvement on rainy conditions on Waymo is limited. While we show results on similar Argoverse settings for consistency and completeness, our harsh settings split results in very few samples (in particular for Argoverse Night). Hence, results become less meaningful. While insightful, this experiment demonstrates the difficulty in obtaining diverse annotated data, further motivating our SelfD approach for development and real-world scalability.  

\boldparagraph{Impact of Initial Training Data} We further investigate the role of the initial training dataset on the pseudo-labeling step and resulting model generalization performance in Table~\ref{tab:diff_init}. Here, we repeat the experiments by replacing nS-Boston with Argoverse and Waymo and test on nS-Singapore (\ie, for direct comparison with Table~\ref{tab:main}). We keep all hyper-parameters fixed. While models trained on different dataset may bias differently the pseudo-labels on the YouTube dataset, overall we find similar trends in performance regardless of the original training dataset. 

\begin{figure*}[!t] 
           \centering
       \includegraphics[trim=0cm 2.2cm 2.4cm 0cm, width=3.3in]{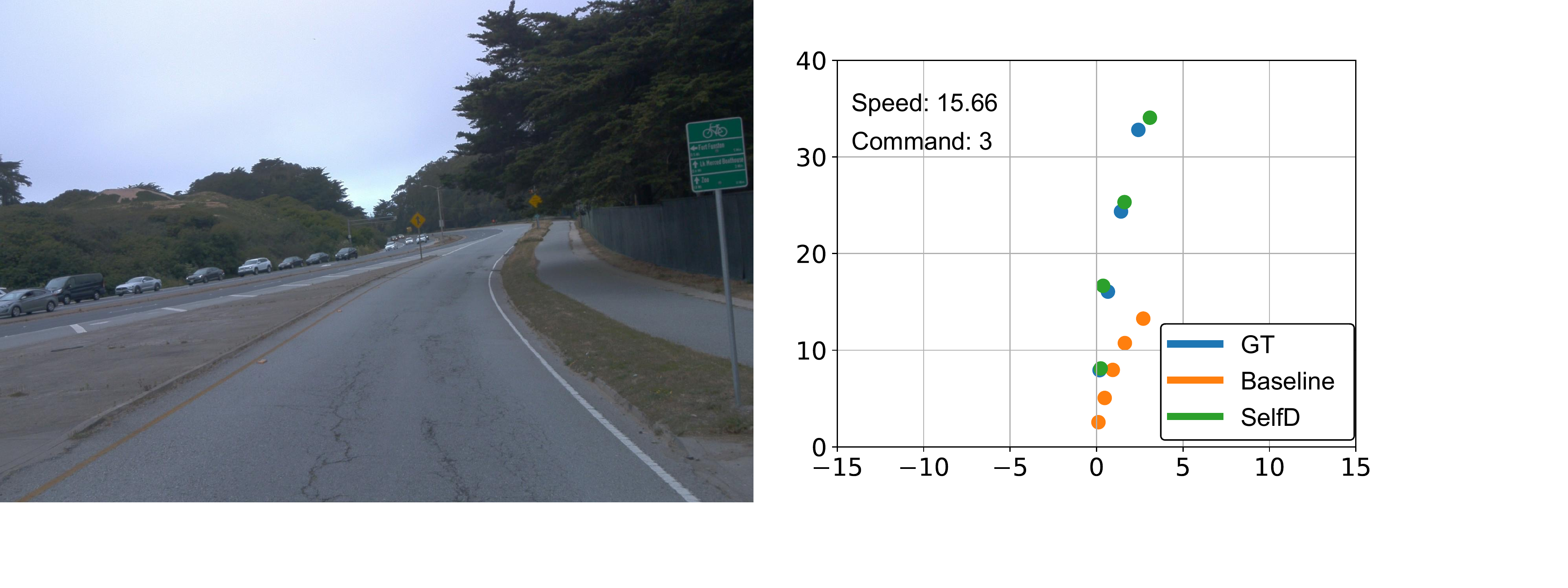} 
       \includegraphics[trim=0cm 2.2cm 2.4cm 0cm, width=3.3in]{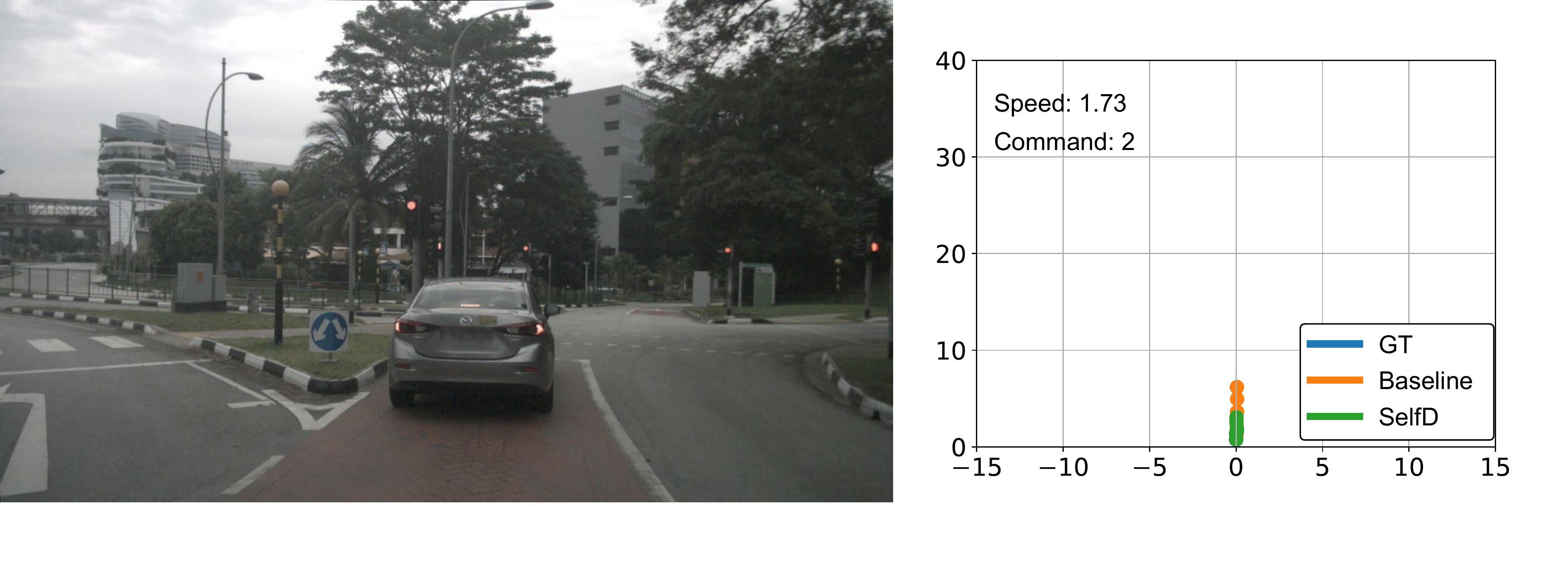}\\ \vspace{0.1cm}
       \includegraphics[trim=0cm 2.2cm 2.4cm 0cm, width=3.3in]{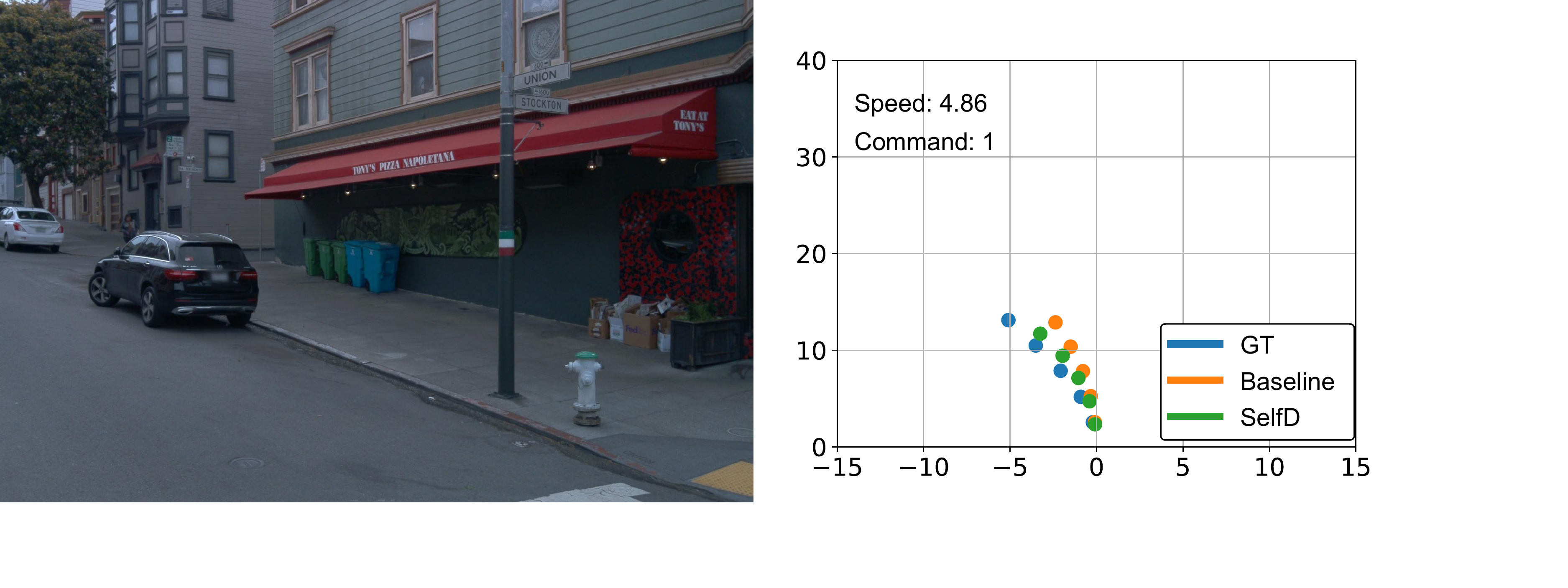} 
       \includegraphics[trim=0cm 2.2cm 2.4cm 0cm, width=3.3in]{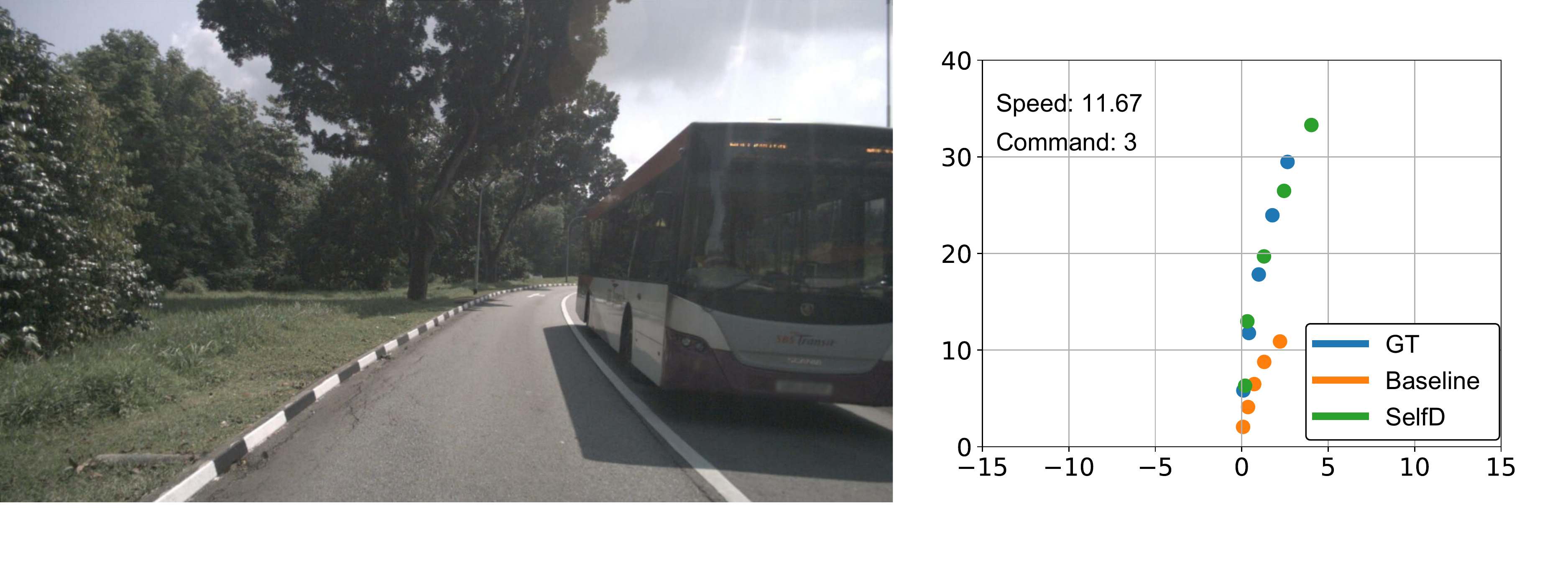}
      \caption{\textbf{Qualitative Results.} We compare predicted waypoints in BEV for the ground truth trajectory, the baseline model (trained on nS-Boston), and SelfD (pre-trained on YouTube, fine-tuned on nS-Boston). Results are shown on Waymo (left) and nS-Singapore (right). The SelfD model (green) is shown to generalize well across unusual scenarios due to the proposed pseudo-labeling mechanism. We also observe improved robustness to rare command and speed inputs. \textbf{Units are meters} for the BEV plan plots.}
   \label{fig:qua}
  \end{figure*}

   \begin{figure}[!t] 
          \centering
      \includegraphics[trim=0cm 0cm 0cm 0cm, width=1.6in]{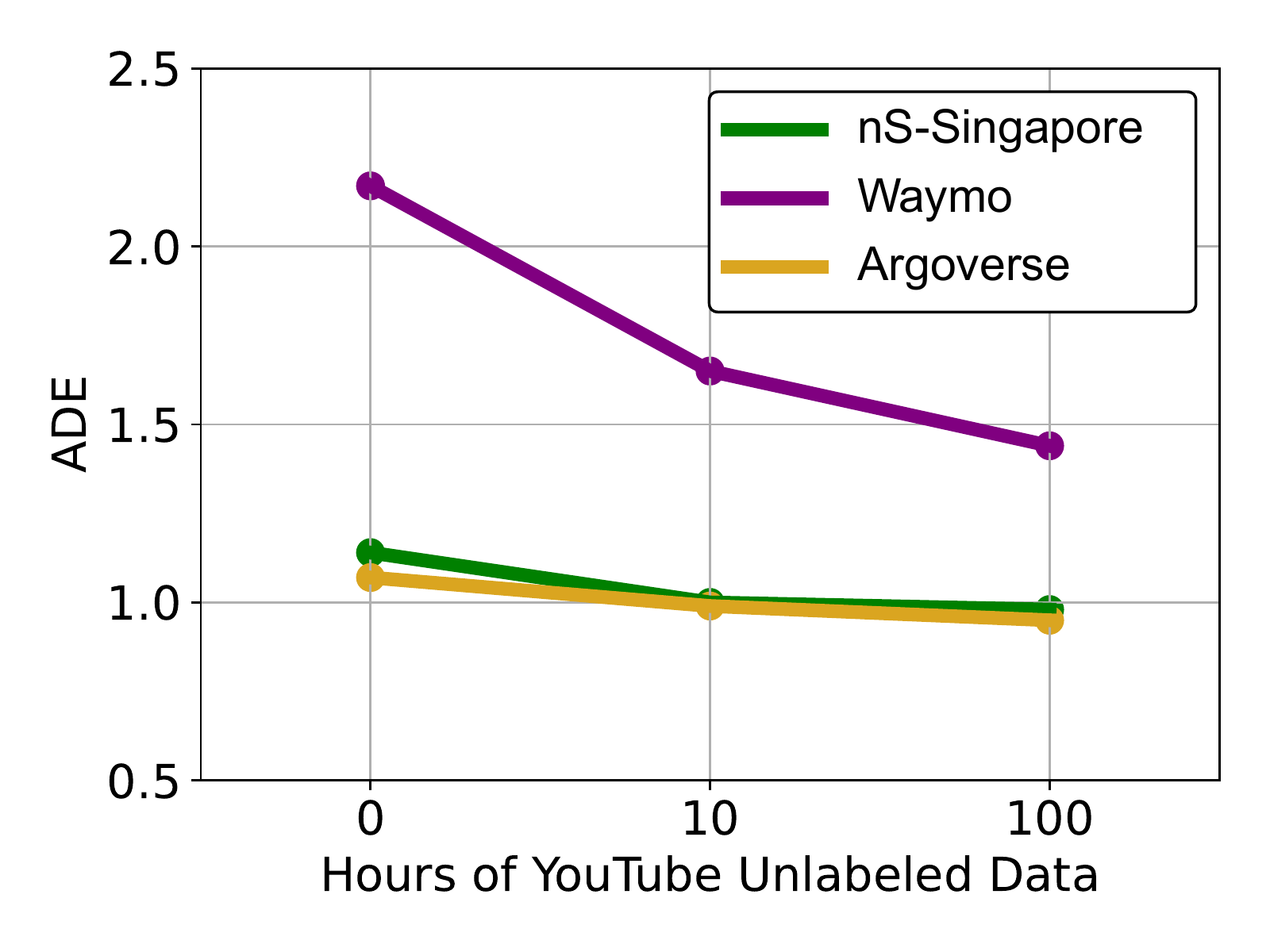} 
      \includegraphics[trim=0cm 0cm 0cm 0cm, width=1.6in]{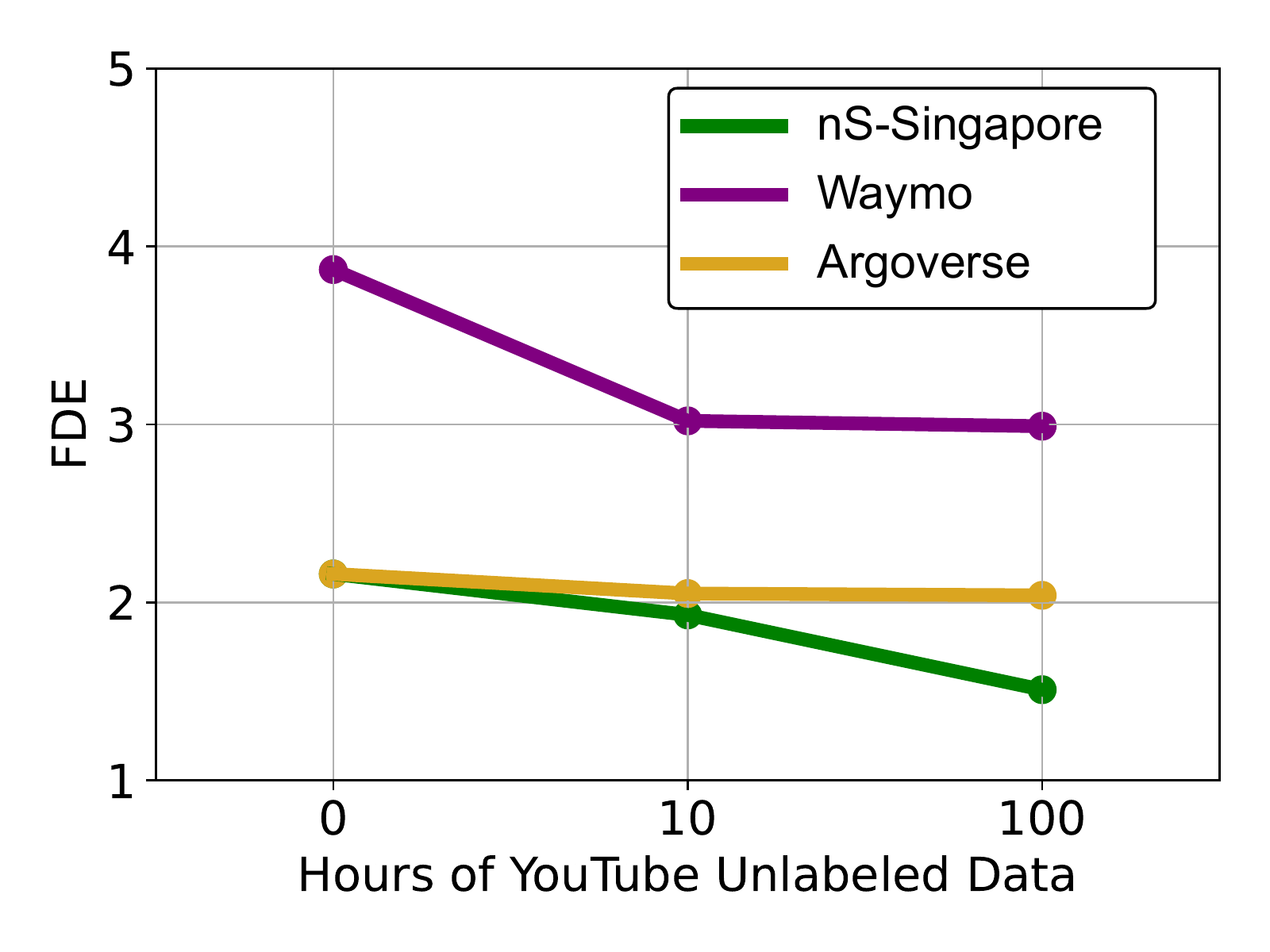}
     \caption{\textbf{Additional Unlabeled Data.} Results when increasing the amount of pseudo-labeled YouTube driving data up to 100 hours. As our proposed self-training pipeline facilitates scalable learning from large amounts of
data, the model can continue to improve with additional YouTube data. 
     }
  \label{fig:hours}
 \end{figure}

\boldparagraph{Closed-Loop Evaluation in CARLA} We show our analysis in CARLA in Table~\ref{tab:carla}. We report both open-loop and closed-loop metrics for completeness. In this experiment, we train our BEV planner using an annotated portion of Town 1, and test on Town 2. We note that while our CIL Baseline benefits from the known fixed perspective transform assumption the BEV planner has no knowledge of camera parameters and must learn it from the data. When training SelfD using pseudo-labeled data from Town 01, we achieve a near doubling in the success rate, from 12$\%$ to $25\%$. This increase in model performance using the most difficult closed-loop evaluation metric highlights the benefits of the proposed approach. We further pseudo-label and use in pre-training an unlabeled dataset from Town 2 (\ie, SelfD - Town 1 \& 2), showing additional gains in driving performance across metrics. This experiment suggests multiple ways in which the proposed approach may be leveraged in real-world generalization and adaptation settings.

\boldparagraph{Additional YouTube Pre-Training Data} While we leverage 10 hours of YouTube data in our ablation analysis, we provide results with additional YouTube data of up to 100 hours. The results, shown in Fig.~\ref{fig:hours}, suggest further gains can be made with larger and more diverse pre-training datasets from the web. As our proposed self-training pipeline facilitates scalable learning from large amounts of data, the model may continue to improve with additional YouTube data. However, this requires further study in the future, \eg, with respect to larger network capacity.

\section{Conclusion}

We envision broad and easily deployable autonomous navigation systems. However, access to resources and data limits the scope of the brittle autonomous systems today. Our SelfD approaches enables to significantly improve an initially trained policy without incurring additional data collection or annotation efforts, \ie, for a new platform, perspective, use-case, or ambient settings. 
Crucially, due to the proposed underlying model architecture, we do not incorporate camera parameters or configuration assumptions into the monocular inference. As SelfD is self-improving, a future direction could be to continue and learn from increasingly larger online datasets beyond what is described in our study. While we emphasized efficient large-scale training in our model development, how to best extend SelfD to more explicitly leverage temporal demonstration data is still an open question which could be studied further in the future. Finally, beyond complex 3D navigation, it would be interesting to explore the applicability of our proposed training framework for learning various embodied tasks from unlabeled web data. 

\boldparagraph{Acknowledgments}
We thank the Red Hat Collaboratory and the BU Center for Information and Systems Engineering for research awards.

%------------------------------------------------------------------------

{\small
\bibliographystyle{ieee_fullname}
\bibliography{egbib}
}

\begin{figure*}[t]
    \centering
    \begin{tabular}{c}
        \includegraphics[trim=1cm 16cm 1cm 7cm,clip,width=6in]{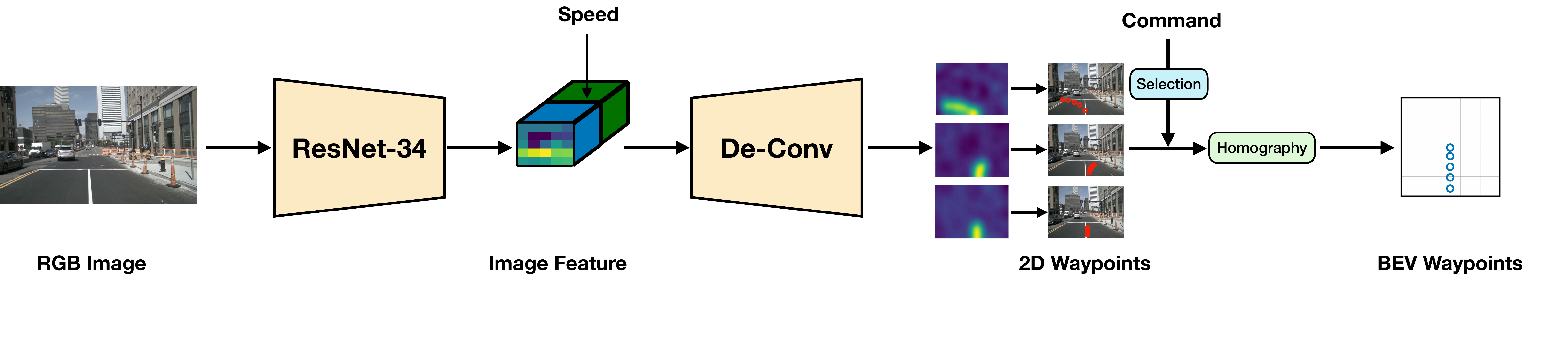} \vspace{-0.7cm}\\  
        (a) CIL Baseline~\cite{lbc} With Known Fixed Homography (Perspective) Transformation.
        \\
        \\
    \includegraphics[trim=1cm 16cm 1cm 7cm,clip,width=6in]{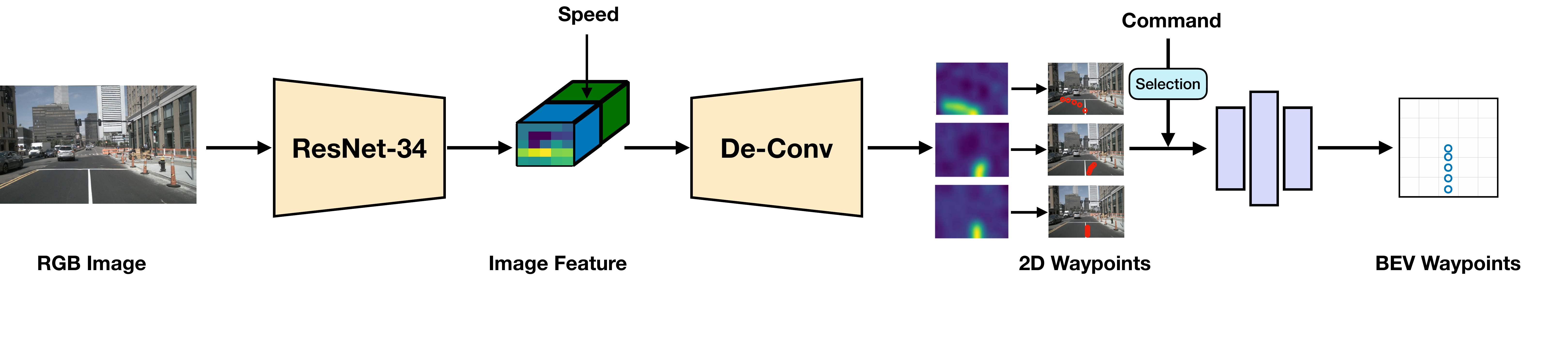}
     \vspace{-0.7cm}\\
        (b) Proposed BEV Planner With Learned Single-Branch Waypoint Transformation.
        \\
        \\
    \includegraphics[trim=1cm 16cm 1cm 7cm,clip,width=6in]{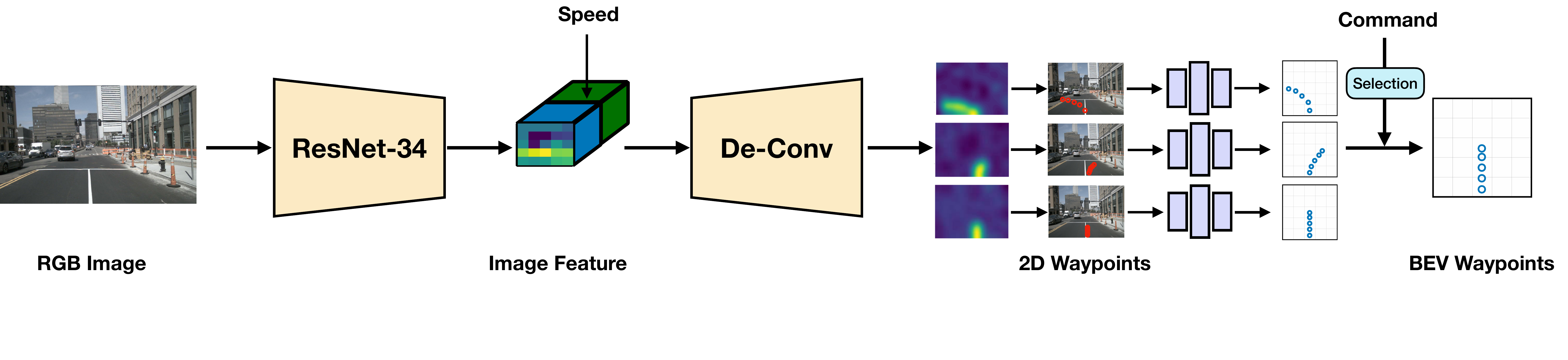} 
        \vspace{-0.7cm}
    \\
        (c) Proposed BEV Planner With Learned Conditional Multi-Branch Waypoint Transformation.
    \end{tabular}
   \vspace{-0.25cm}
    \caption{\textbf{BEV Planner Architecture.} To handle viewpoint and layout diversity, our proposed model does not assume a fixed known BEV perspective transform. A multi-branch projection model (c) provides the best results with 1.14 ADE on NS-Singapore in contrast to 1.86 ADE and 1.21 ADE for (a) and (b), respectively. SelfD employs the conditional waypoint projection module (c).   
    }
    \label{fig:arch}
   \vspace{-0.3cm}
\end{figure*}

\begin{table*}[t]
\caption{\textbf{SelfD with Different Initial Data.} We analyze the impact of the initial pre-training dataset to find consistent trends and improvements in model generalization. 
}
\label{tab:diff_init_complete}
\tablestyle{8pt}{1.1}
\begin{tabular}{l l|c c | c c | c c |c c}
\toprule
\multirow{2}{*}{Initial Dataset} & {\multirow{2}{*}{Method }} & \multicolumn{2}{c|}{nS-Boston} & \multicolumn{2}{c|}{nS-Singapore} & \multicolumn{2}{c|}{Argoverse} & \multicolumn{2}{c}{Waymo} \\
& & ADE & FDE & ADE & FDE   & ADE & FDE & ADE & FDE \\
 \midrule
\multirow{3}{*}{\begin{tabular}{l}nS-Singapore
                \end{tabular}}
&BEV Planner       & 0.79 & 1.50 & - & - & 1.04 & 2.07 & 1.38 & 2.52 \\
&SelfD (Pre-Trained) & 0.95 & 1.79 & - & - & 1.17 & 2.28 & 1.75 & 3.12 \\
&SelfD (Fine-Tuned)  & \bf{0.74} & \bf{1.47} & - & - & \bf{1.00} & \bf{2.07} & \bf{1.29} & \bf{2.45} \\
 \midrule
 \multirow{3}{*}{\begin{tabular}{l}Argoverse
                \end{tabular}}
&BEV Planner        & 1.17 & 2.11 & 1.42 & 2.58 & - & - & 2.16 & 3.78   \\
&SelfD (Pre-Trained)  & 1.17 & 2.10 & 1.38 & 2.52 & - & - & 2.20 & 3.86\\
&SelfD (Fine-Tuned)   & \bf{0.85} & \bf{1.60} & \bf{1.18} & \bf{2.23} & - & - & \bf{1.71} & \bf{3.13}  \\

 \midrule
 \multirow{3}{*}{\begin{tabular}{l}Waymo
                \end{tabular}}
&BEV Planner         & 0.78 & \bf{1.49} & 1.08 & 2.03 & 1.04 & 2.06 & - & - \\
&SelfD (Pre-Trained)   & 0.87 & 1.63 & 1.13 & 2.14 & 1.11 & 2.19 & - & -\\
&SelfD (Fine-Tuned)    & \bf {0.77} & 1.52 & \bf{0.91} & \bf{1.81} & \bf{0.99} & \bf{2.06} & - & -  \\
\bottomrule
\end{tabular}
\end{table*}

\section*{Appendix A. Implementation and Training}
This section details the BEV plan network architecture and training procedure. 

\subsection*{A.1. Network Architecture}
\boldparagraph{Learned BEV Plan Transformation}
Fig.~\ref{fig:arch} contrasts our proposed BEV plan network architecture with two baselines. The first baseline is the implementation of the waypoint-based CIL~\cite{lbc}, as depicted in Fig.~\ref{fig:arch}(a). This common approach utilizes a conditional command to select a set of 2D waypoints in the image-plane. The image-plane waypoints can then be projected to the BEV using a known perspective transformation. Based on our experiments on nS-Singapore, we find the assumption of a fixed perspective transform (which may often hold in CARLA) to degrade BEV projection quality. By learning the image-to-BEV projection as shown in Fig.~\ref{fig:arch}(b), we find
a significant reduction in ADE on the nS-Singapore test set, from $1.86$ for the image-plane CIL baseline~\cite{lbc} to $1.21$. Here, the overall network structure is kept identical with the exception of a learned projection implemented with 3 Fully-Connected (FC) layers (first two with Relu and Dropout). However, while this architecture does not depend on a pre-calibrated homography matrix, we find this single learned projection branch to be insufficient for the BEV prediction task in many real-world scenarios. In particular, a single branch network may overfit to the main mode of the training data, \ie, of driving straight, addressed next.

\boldparagraph{Conditional Multi-Branch Transformation} 
To accurately project waypoints across various maneuvers, we propose to leverage conditional projection branches. As shown in Fig.~\ref{fig:arch}(c), by selecting the BEV waypoints based on the conditional command, the model can learn more robust projection for the less frequent turn events and their BEV waypoint structure. Our SelfD experiments therefore utilize the last network architecture in Fig.~\ref{fig:arch}(c). All three architectures take RGB images, speed, and conditional command as input. The image is processed through a ResNet-34 backbone. Next, the image features are concatenated with the measured speed and forwarded to deconvolution layers for computing image-plane waypoint heatmaps followed by a spatial softmax to determine final waypoints. 
We next discuss the training procedure for the network.

\boldparagraph{Training Details} 
Throughout the experiments, we initialize our policy model from ImageNet~\cite{deng2009imagenet}. We optimize the network using Adam~\cite{Kingma2015ICLR} with a batch size of $96$ for $128$ epochs. The learning rate is set to \num{1e-3}. 
\begin{table*}[!t]
\tablestyle{13pt}{1.1}
\caption{\textbf{Additional Iterations.} An additional iteration of pseudo-labeling and re-training provides limited benefits. 
}
\label{tab:iter}
\begin{tabular}{l l|c c | c c |c c}
\toprule
\multicolumn{2}{c|}{\multirow{2}{*}{Method }} & \multicolumn{2}{c|}{nS-Singapore} & \multicolumn{2}{c|}{Argoverse} & \multicolumn{2}{c}{Waymo} \\
& & ADE & FDE & ADE & FDE   & ADE & FDE  \\
 \midrule
\multicolumn{2}{c|}{BEV Planner (nS-Boston)}  & 1.14 & 2.16 &1.07 & 2.16  & 2.17 & 3.87 \\
 \midrule
 \multirow{2}{*}{\begin{tabular}{l}Iteration 1
                \end{tabular}}
&SelfD (Pre-Trained)  & 1.19 & 2.28 & 1.13 & 2.28 & 2.24 & 4.01  \\
&SelfD (Fine-Tuned)  & \textbf{1.00} & 1.93 & \textbf{0.99} & \textbf{2.05} & \textbf{1.65} & \textbf{3.02} \\
\midrule
\multirow{2}{*}{\begin{tabular}{l}Iteration 2
                \end{tabular}}
&SelfD (Pre-Trained)  & \textbf{1.00} & \textbf{1.92} & 1.04 & 2.10 & 1.77 & 3.17  \\
&SelfD (Fine-Tuned)  & 1.07 & 2.06 & 1.04 & 2.12 & 2.18 & 3.87\\
\bottomrule
\end{tabular}
\end{table*} 

\begin{table}[!t]
      \caption{\textbf{Harsh Conditions Data Statistics.} Number of driving episodes within each harsh evaluation condition. Given the difficulty in obtaining diverse data with annotations, this split is restrictive for certain scenes and datasets. }
     \label{tab:dataset}
     \centering
     \noindent\adjustbox{max width=\columnwidth}{
     \begin{tabular}{l| c | c |c }
     \toprule
      Dataset & Total Scenes & Night Scenes & Rainy Scenes \\
      \midrule
     nS-Singapore & 383 & 99 & 0\\
     \midrule
     Argoverse & 66 & 3 & 1\\
     \midrule
     Waymo & 150 & 28 & 9\\
     \bottomrule
     \end{tabular}}

 \end{table}
\section*{Appendix B. Additional Experimental Results}

In this section, we first analyze the impact of different initial training datasets on self-training and generalization performance. Secondly, we outline dataset statistics for the analysis in the main paper regarding harsh evaluation settings. Thirdly, we analyze the benefits of an additional iteration of pseudo-labeling with the proposed approach. Finally, we discuss additional qualitative results for the data augmentation strategy and model inference.

\boldparagraph{Impact of Initial Training Data} 
We further investigate the impact of different initial training data on the final policy. Results obtained by initially training with each of the three datasets, nS-Singapore, Argoverse and Waymo, are shown in Table ~\ref{tab:diff_init_complete}. Overall, we find that SelfD improves driving metrics across the settings and datasets regardless of the initial training data. 

\boldparagraph{Additional Iterations}
As shown in Table~\ref{tab:iter}, while the fine-tuned SelfD model could be used to repeat the pseudo-labeling process with a more robust model, we do not find benefits from such additional iterations. Although performance may go up for the pre-trained model due to the less noisy trajectories, this does not outperform prior results and no further gains are observed after an additional fine-tuning. 

\boldparagraph{Generalization to Harsh Settings} 
To gain insights into the underlying performance of the proposed SelfD approach, the main paper analyzes additional testing splits, controlled for weather and ambient conditions. The proposed approach generally improves generalization, \eg, from driving in nS-Boston at daytime to nS-Singapore driving at nighttime. To contextualize the results, we further provide the resulting statistics for this split in Table~\ref{tab:dataset}. nS-Singapore has 99 night scenes without any rainy scenes. Waymo has 150 total scenes, 28 of which are night scenes and 9 are rainy scenes. Argoverse has only three night scenes and one rainy scenes out of 66 total scenes. Thus, while less representative, we also show performance on Argoverse subsets for completeness. We also leverage this results in order to highlight the difficulty in collecting highly diverse and annotated real-world data. Another interesting observation is the minimal performance gains on rainy conditions on both Argoverse and Waymo. Upon further inspection, we realized many of the training scenes in nS-Boston already have cloudy to light rain weather, thus benefiting less from additional unlabeled data within such settings. As the baseline experiences such conditions in training it may already perform well within such conditions without additional pseudo-labeled data. Nonetheless, given the need for larger and more diverse evaluation sets across all settings in Table~\ref{tab:dataset}, this requires further study by future work over more representative datasets.

\begin{figure*}[!t] %!t
           \centering
       \includegraphics[trim=0cm 8.5cm 2.5cm 0cm, width=3.3in]{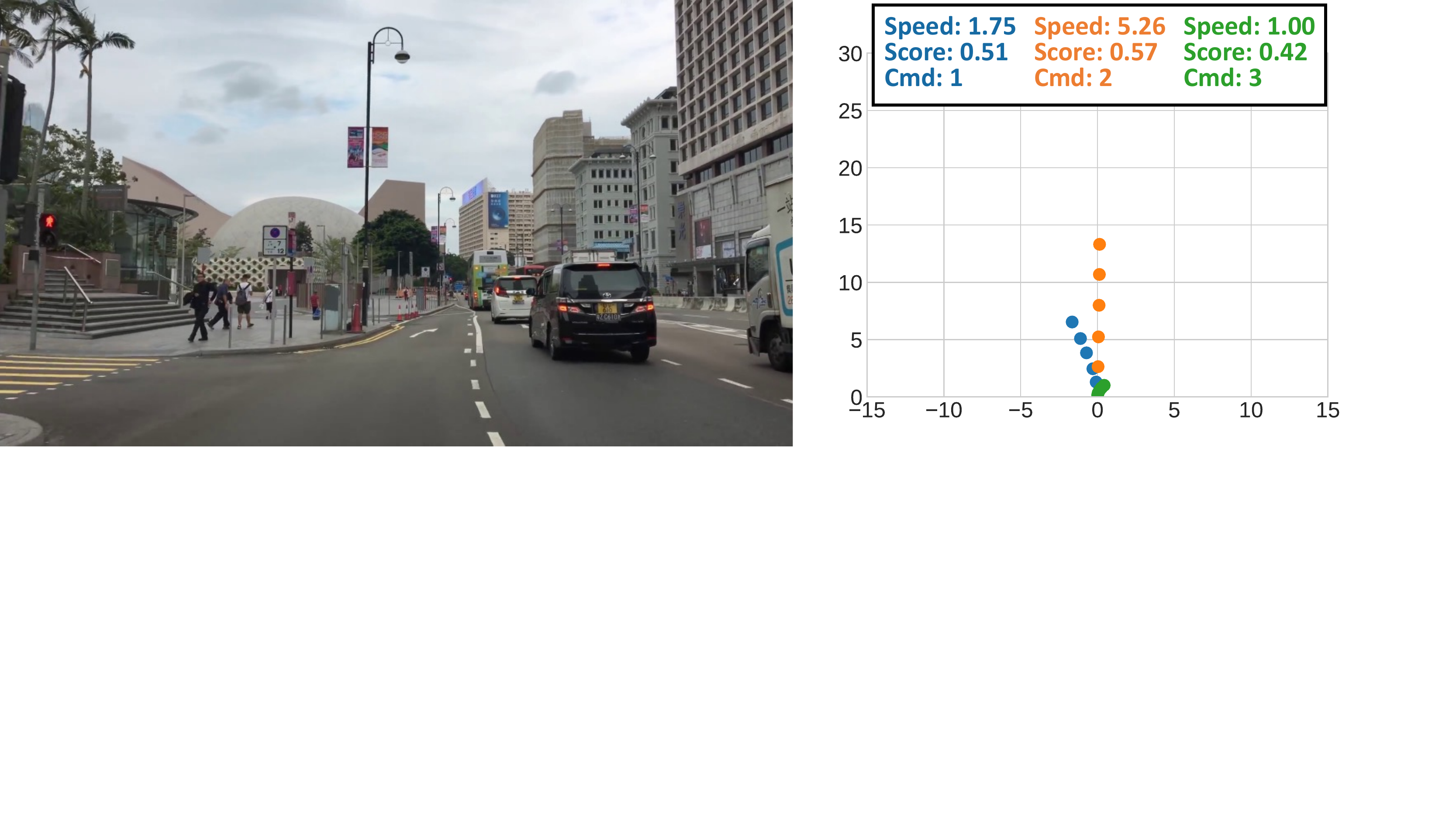} 
       \includegraphics[trim=0cm 8.5cm 2.5cm 0cm, width=3.3in]{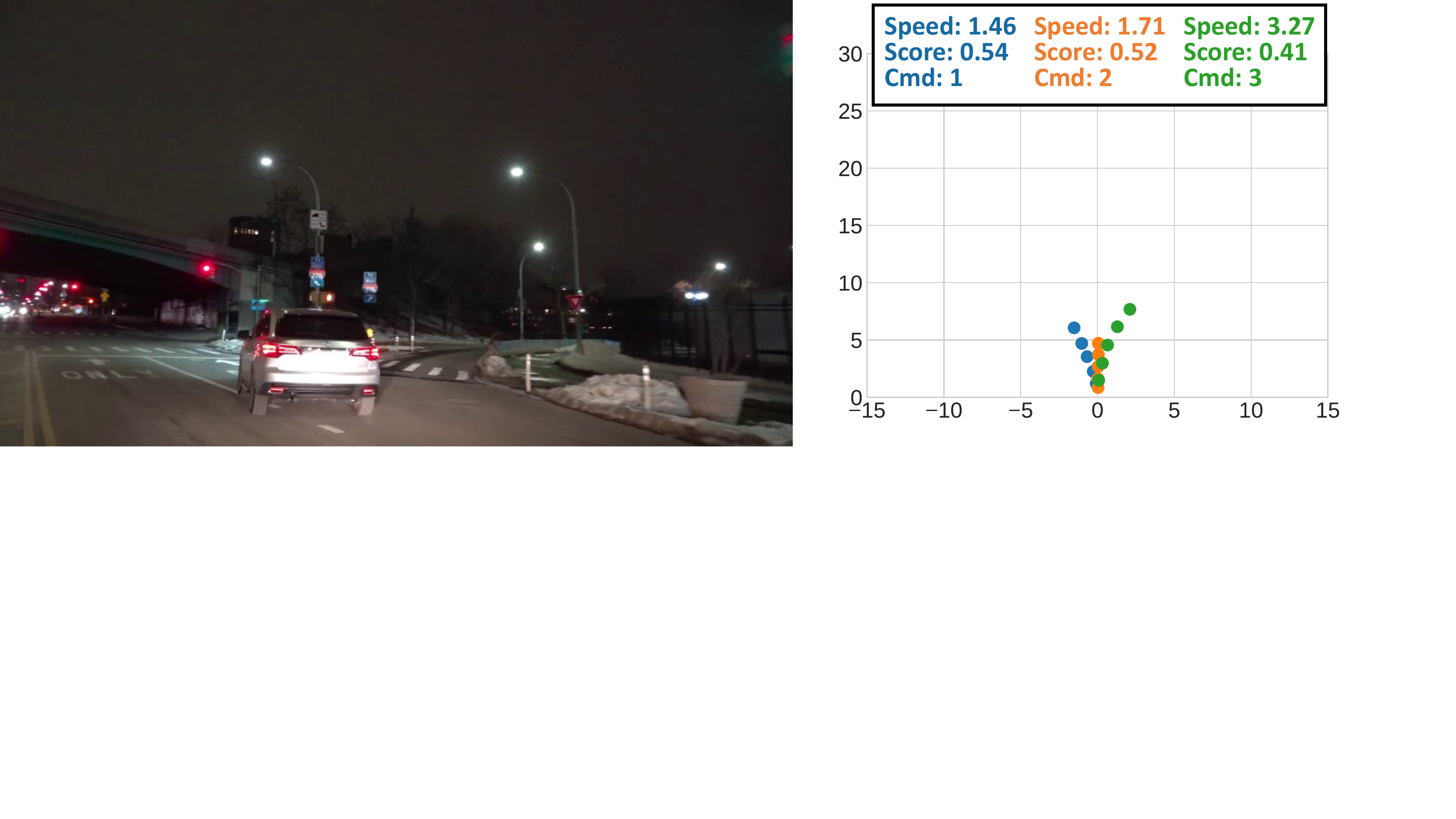}\\ 
       \includegraphics[trim=0cm 8.5cm 2.5cm 0cm, width=3.3in]{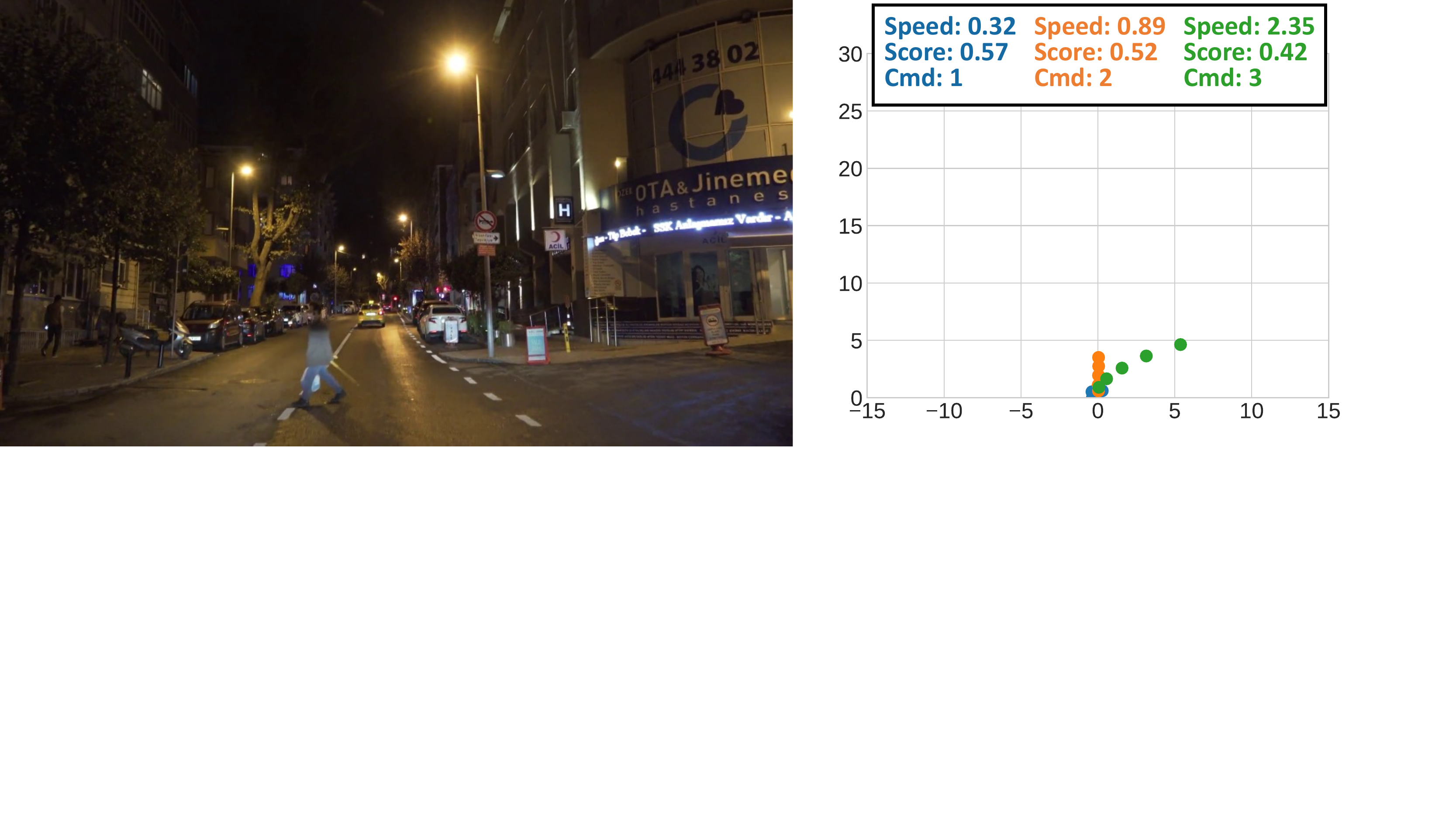} 
       \includegraphics[trim=0cm 8.5cm 2.5cm 0cm, width=3.3in]{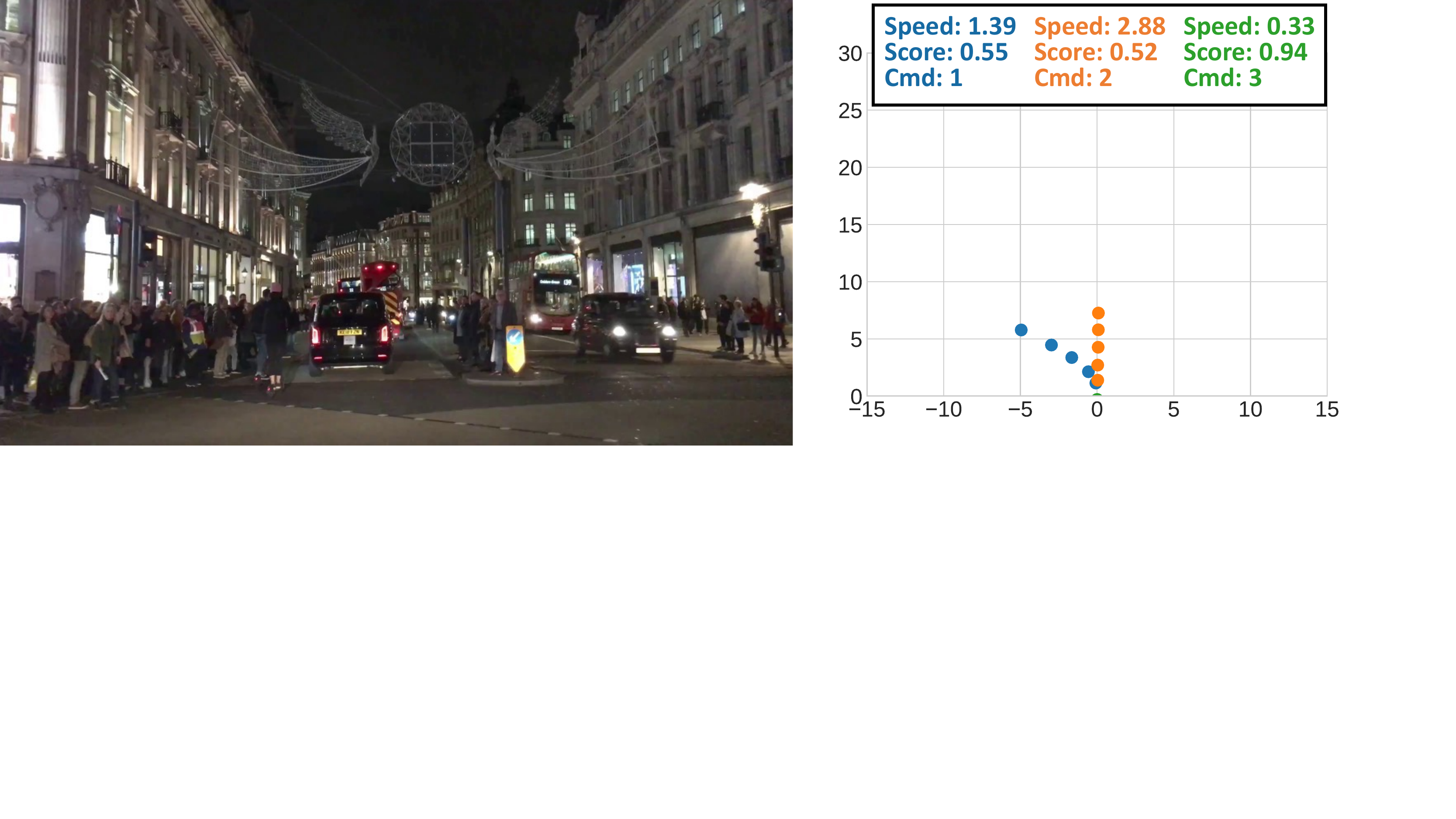}\\
       \includegraphics[trim=0cm 8.5cm 2.5cm 0cm, width=3.3in]{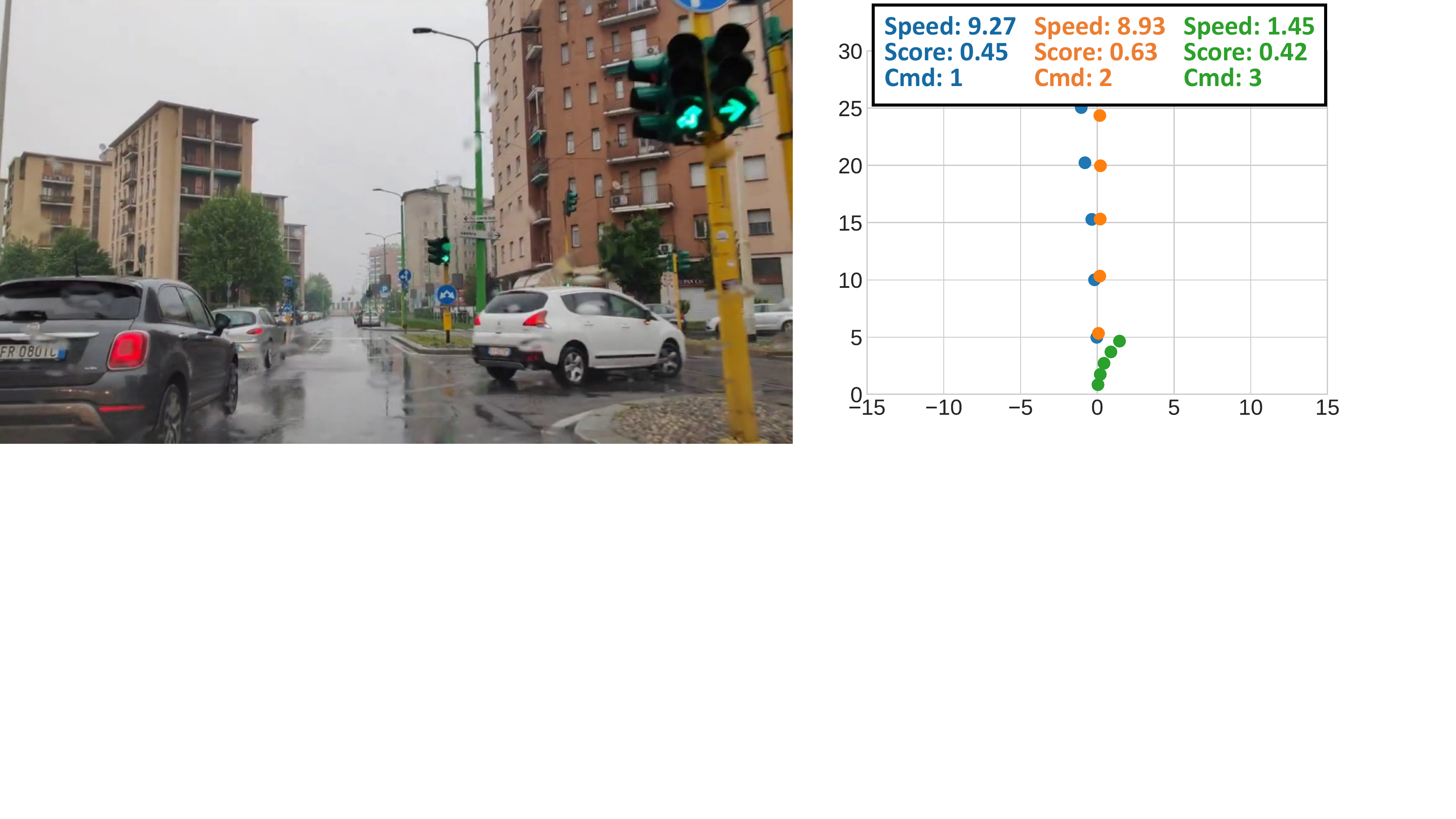} 
       \includegraphics[trim=0cm 8.5cm 2.5cm 0cm, width=3.3in]{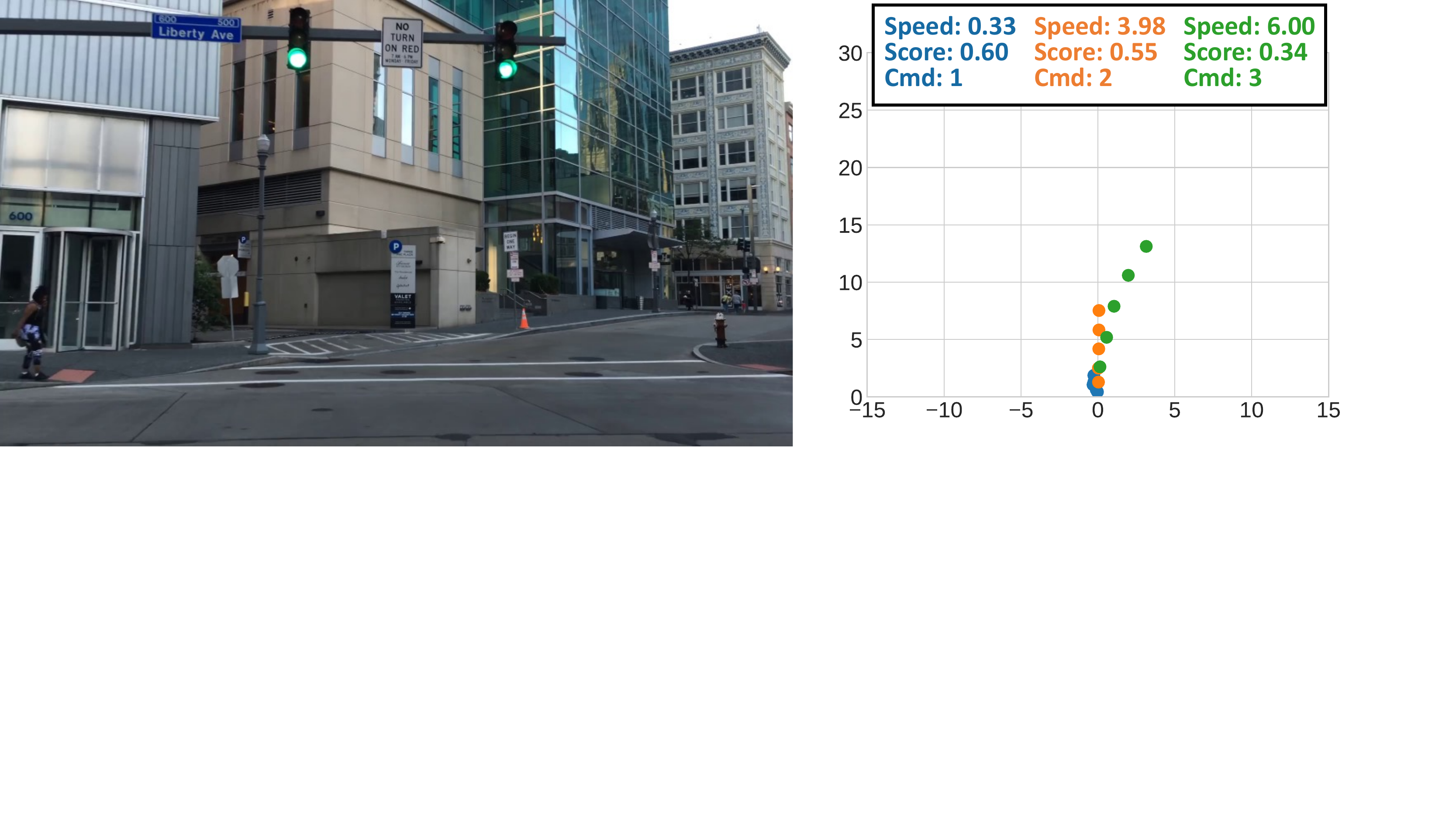}\\
       \includegraphics[trim=0cm 8.5cm 2.5cm 0cm, width=3.3in]{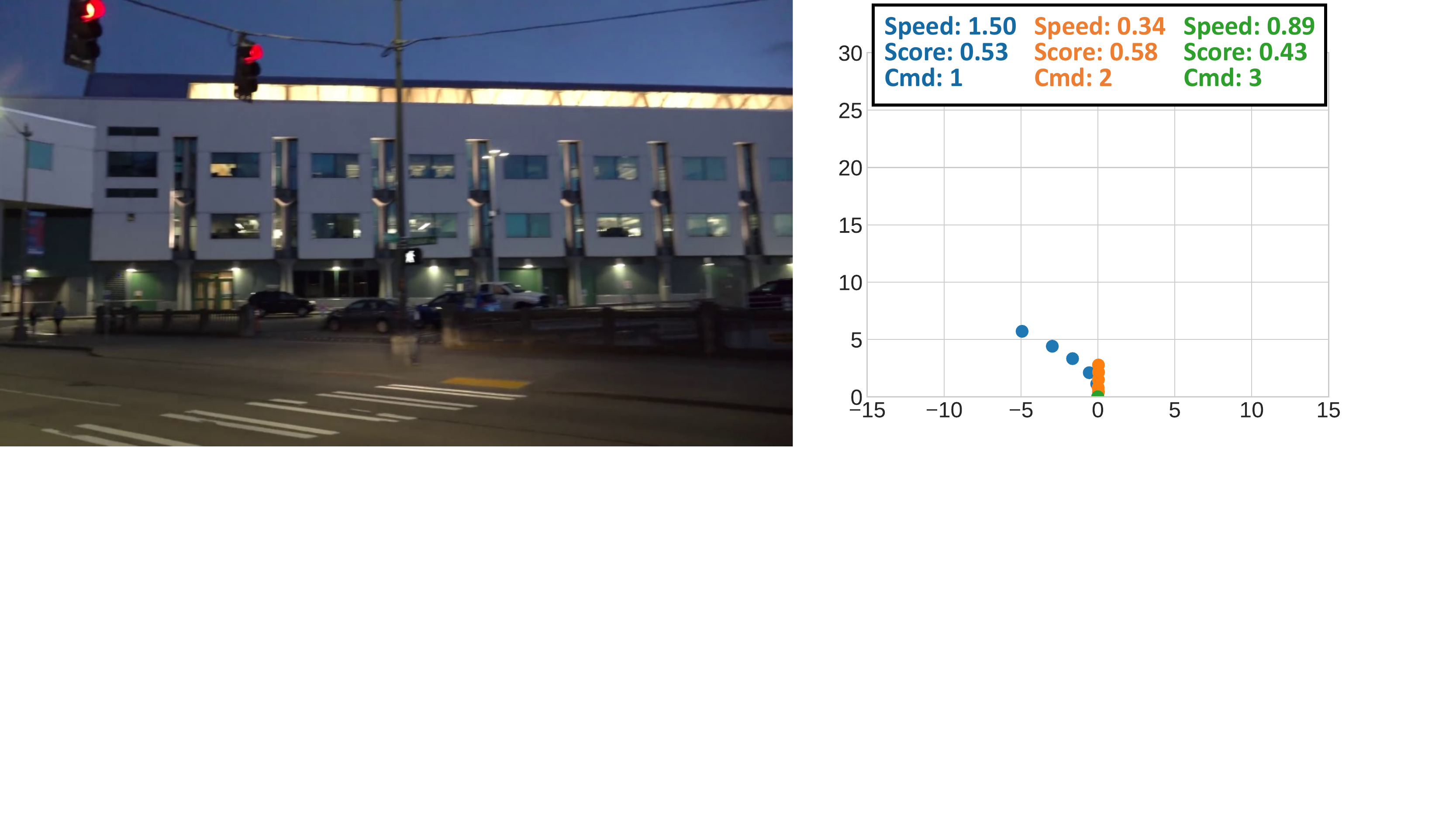} 
       \includegraphics[trim=0cm 8.5cm 2.5cm 0cm, width=3.3in]{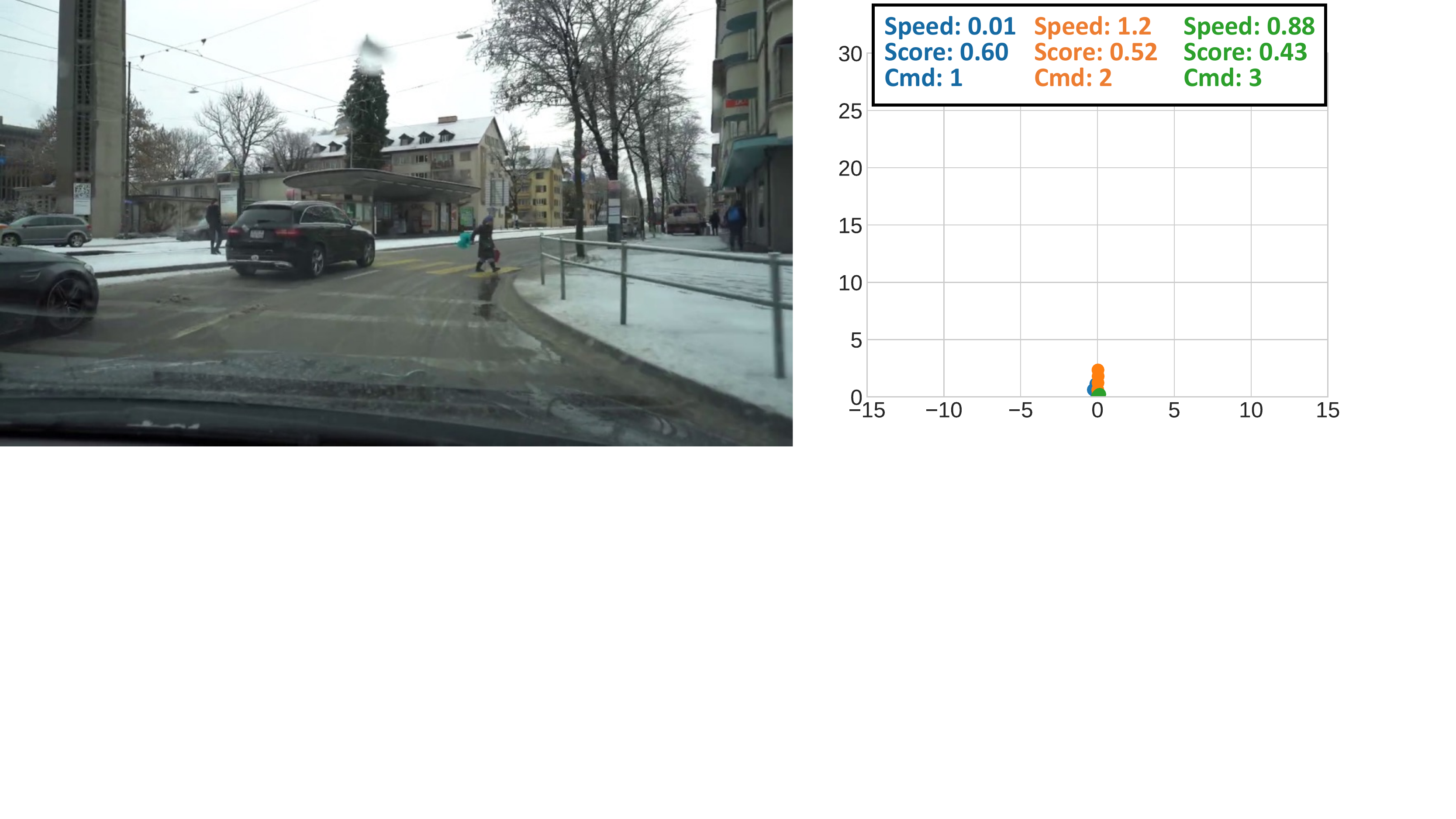}\\
       
      \caption{\textbf{`What If' Pseudo-Labeling.} Additional qualitative examples of `What if' pseudo-labeling augmentation.}
   \label{fig:whatif}
 % \vspace{-0.15in}
  \end{figure*}

\boldparagraph{`What If' Pseudo-Labeling}
To provide details into the proposed data augmentation step, Fig.~\ref{fig:whatif} depicts additional results when sampling from the planner on the unlabeled online scenes. We visualize three pseudo-labels for each driving samples for three different conditional commands and randomly sampled speed. While traditionally each imitation learning data sample is aligned with a single ground truth action label, the `what if' augmentation process provides additional supervision through many plausible samples.

\begin{figure*}[!t] %!t
           \centering
       \includegraphics[trim=0cm 7cm 1.5cm 0cm, width=3.3in]{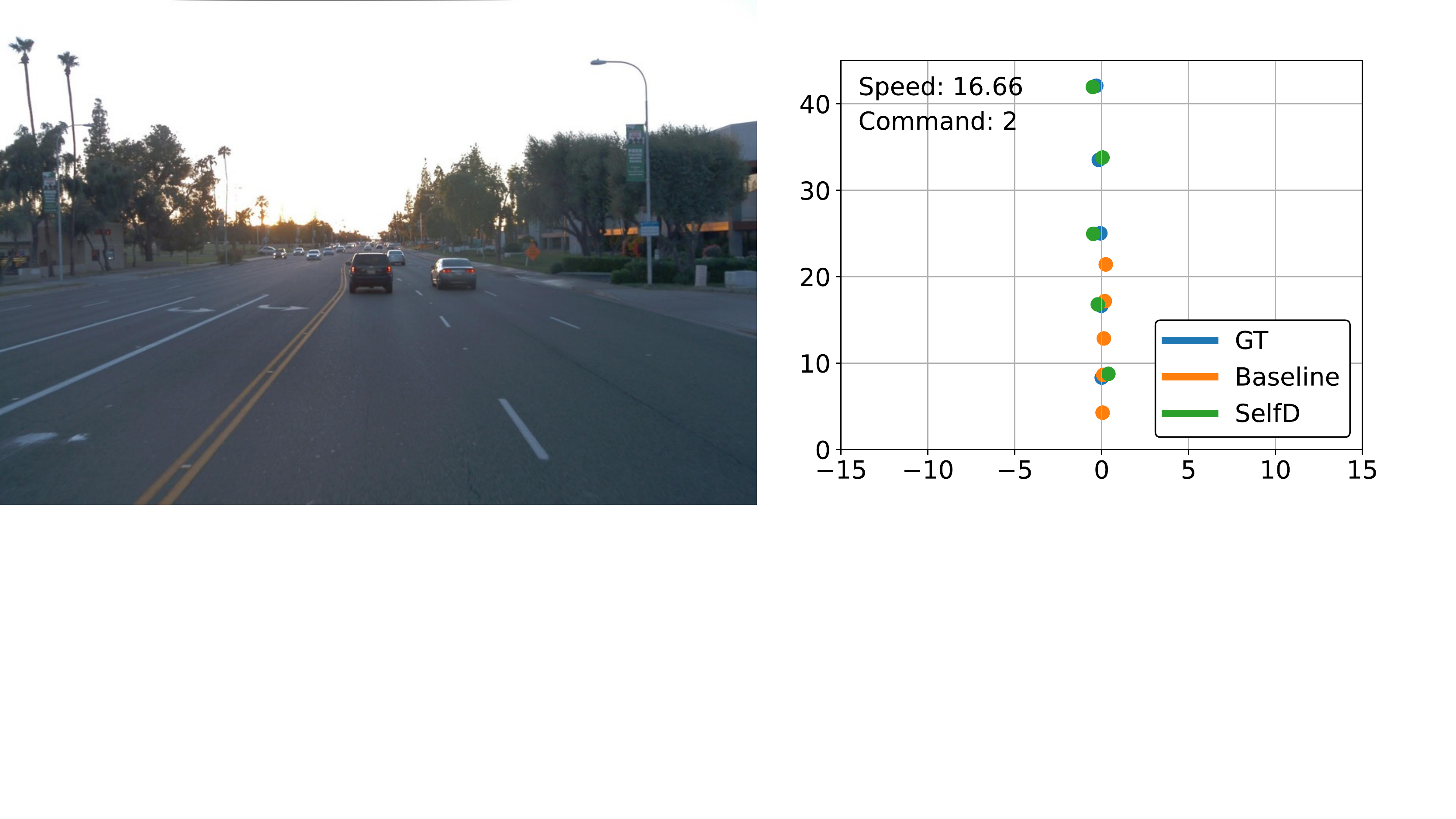} 
       \includegraphics[trim=0cm 7cm 1.5cm 0cm, width=3.3in]{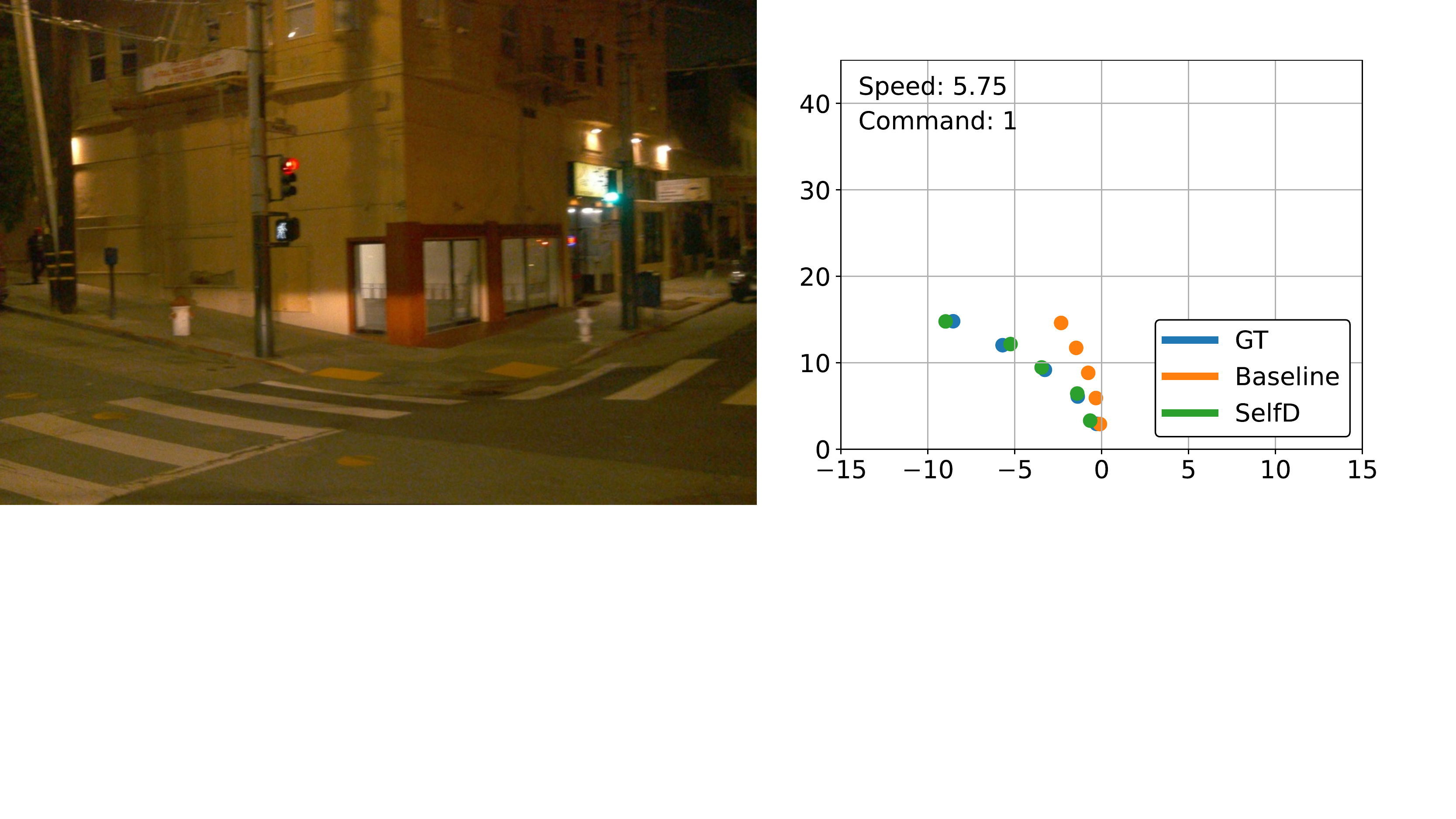}\\ 
       \includegraphics[trim=0cm 7cm 1.5cm 0cm, width=3.3in]{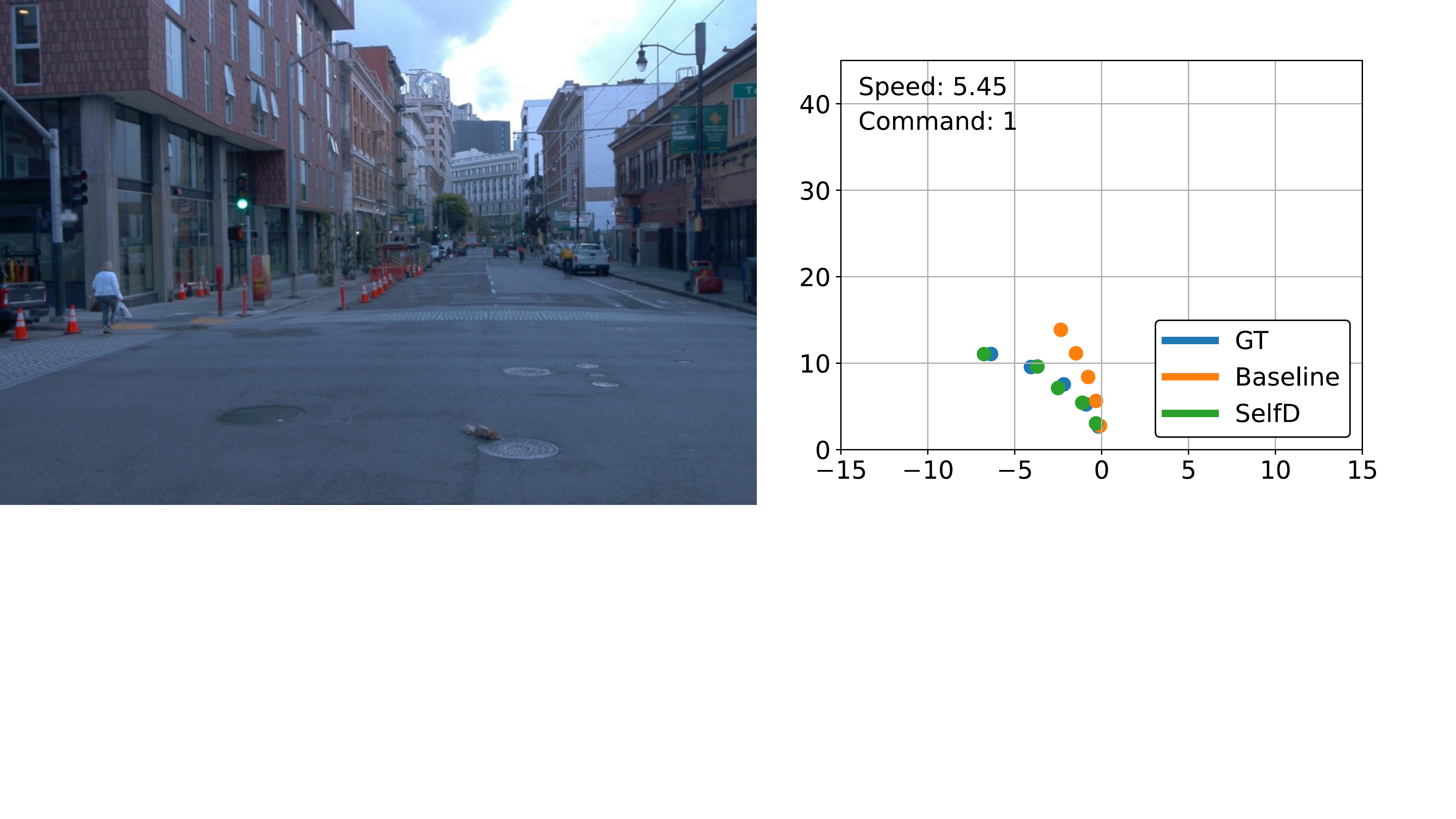} 
       \includegraphics[trim=0cm 7cm 1.5cm 0cm, width=3.3in]{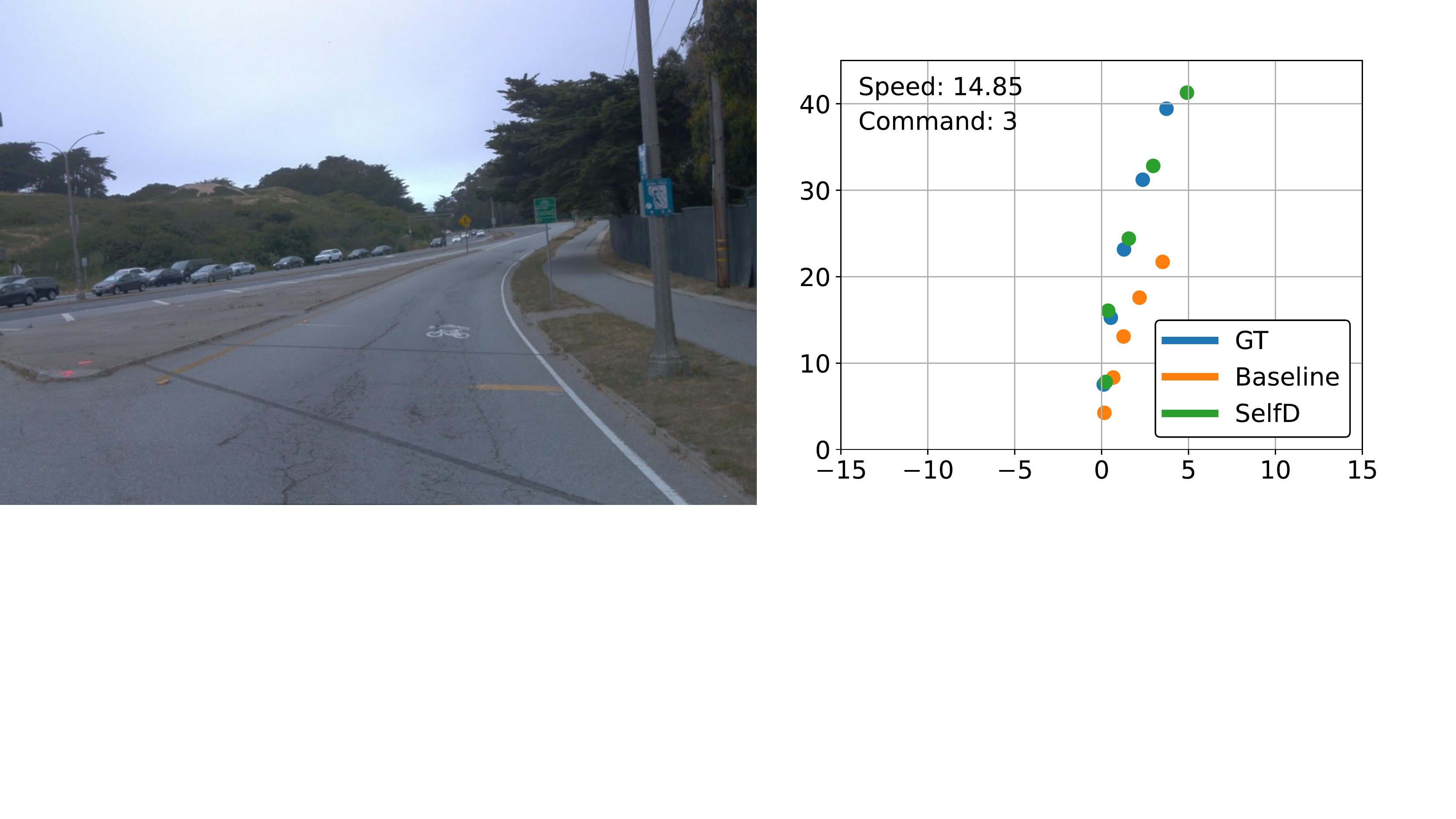}\\
       \includegraphics[trim=0cm 7.3cm 1.5cm 0cm, width=3.3in]{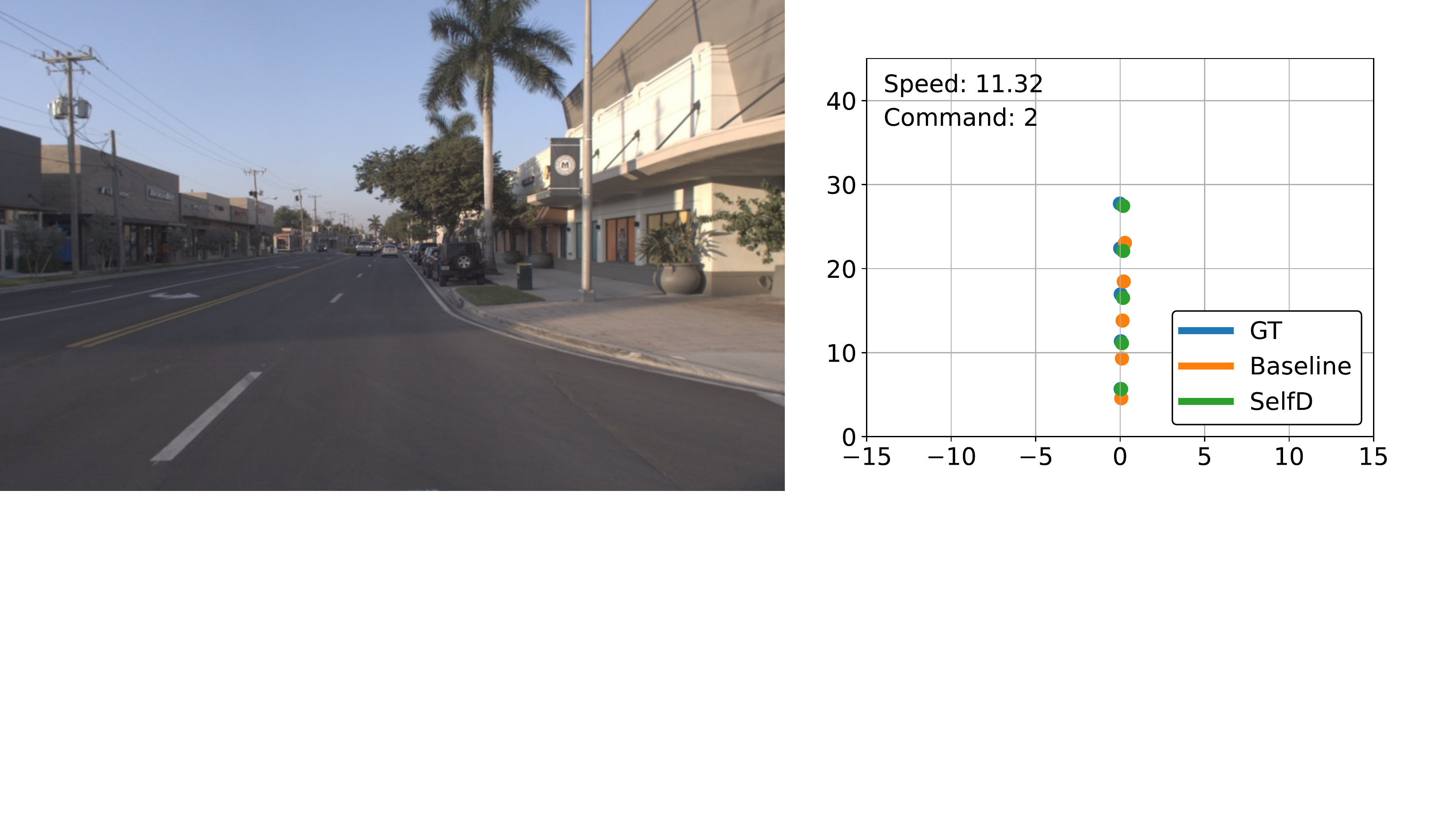} 
       \includegraphics[trim=0cm 7.3cm 1.5cm 0cm, width=3.3in]{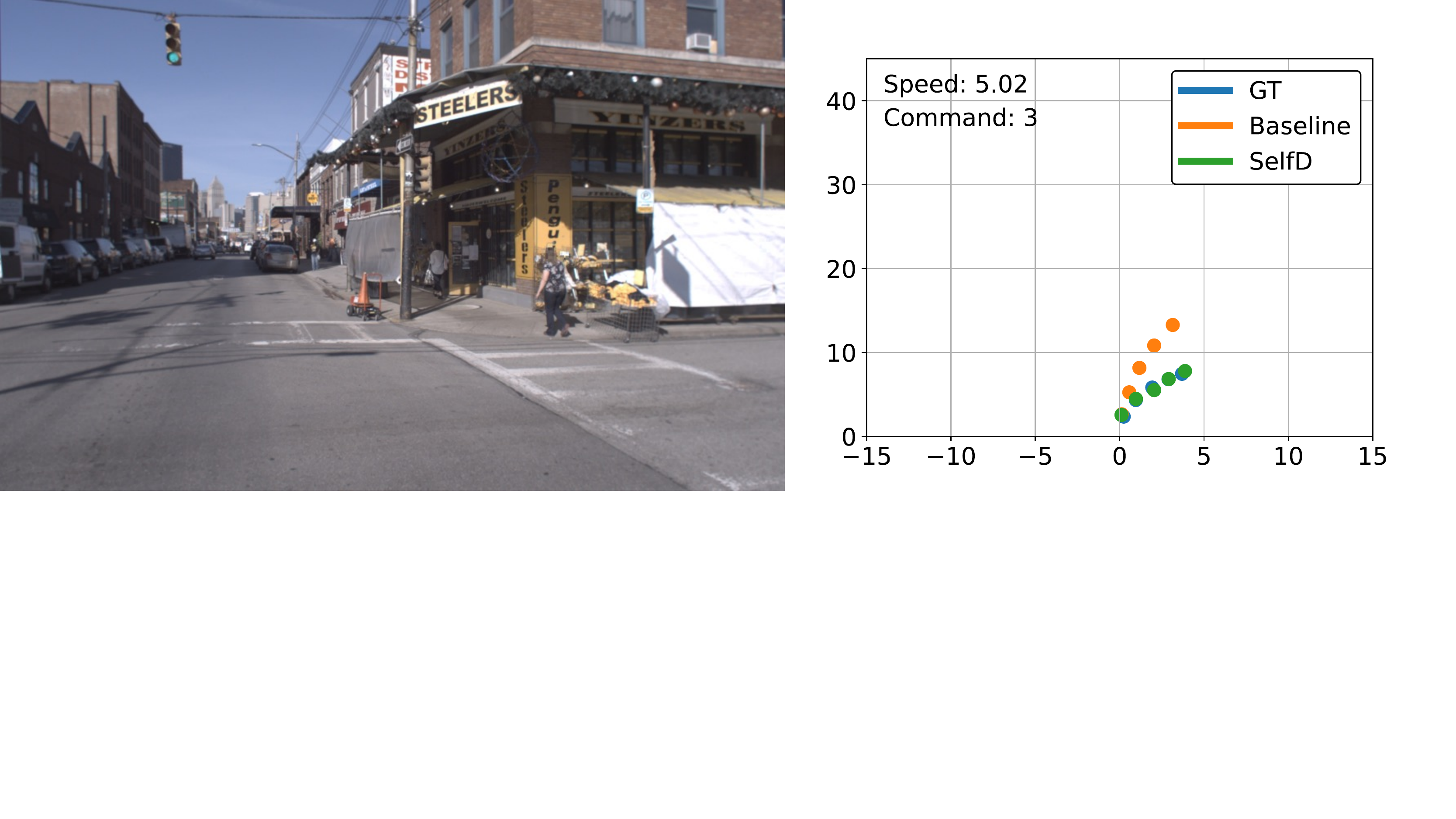}\\
       \includegraphics[trim=0cm 7.3cm 1.5cm 0cm, width=3.3in]{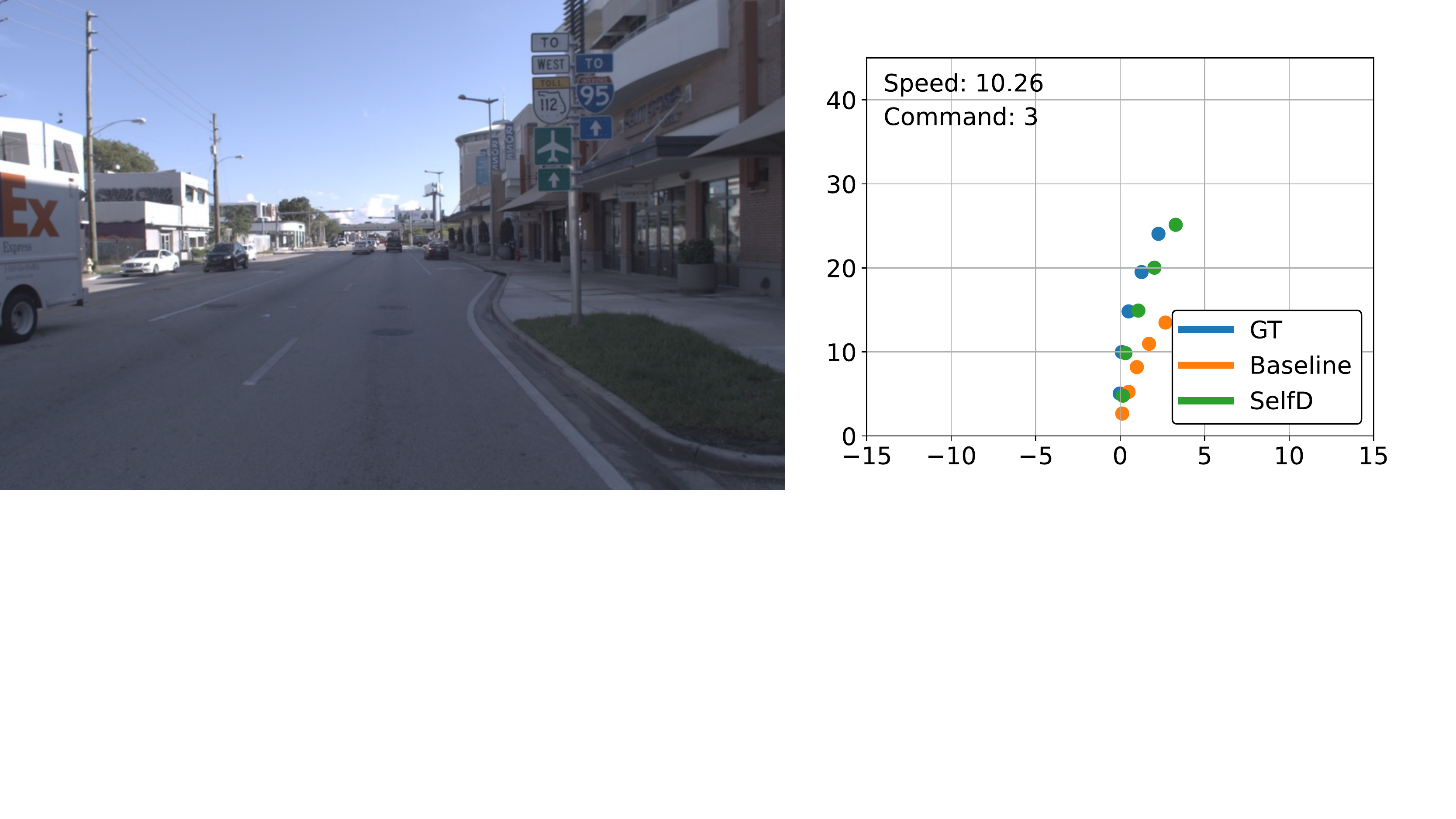} 
       \includegraphics[trim=0cm 7.3cm 1.5cm 0cm, width=3.3in]{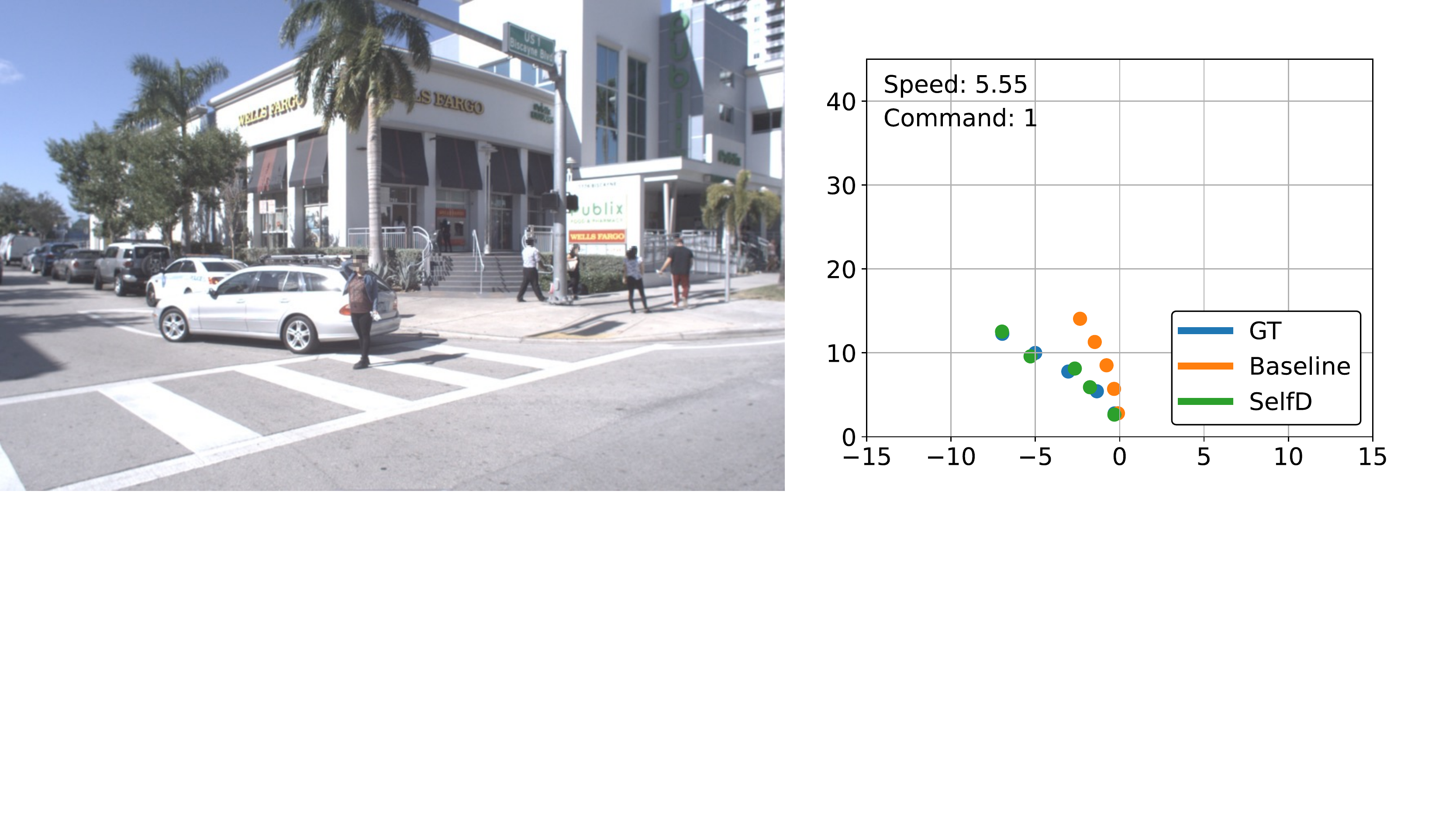}\\
       \includegraphics[trim=0cm 8cm 1.5cm 0cm, width=3.3in]{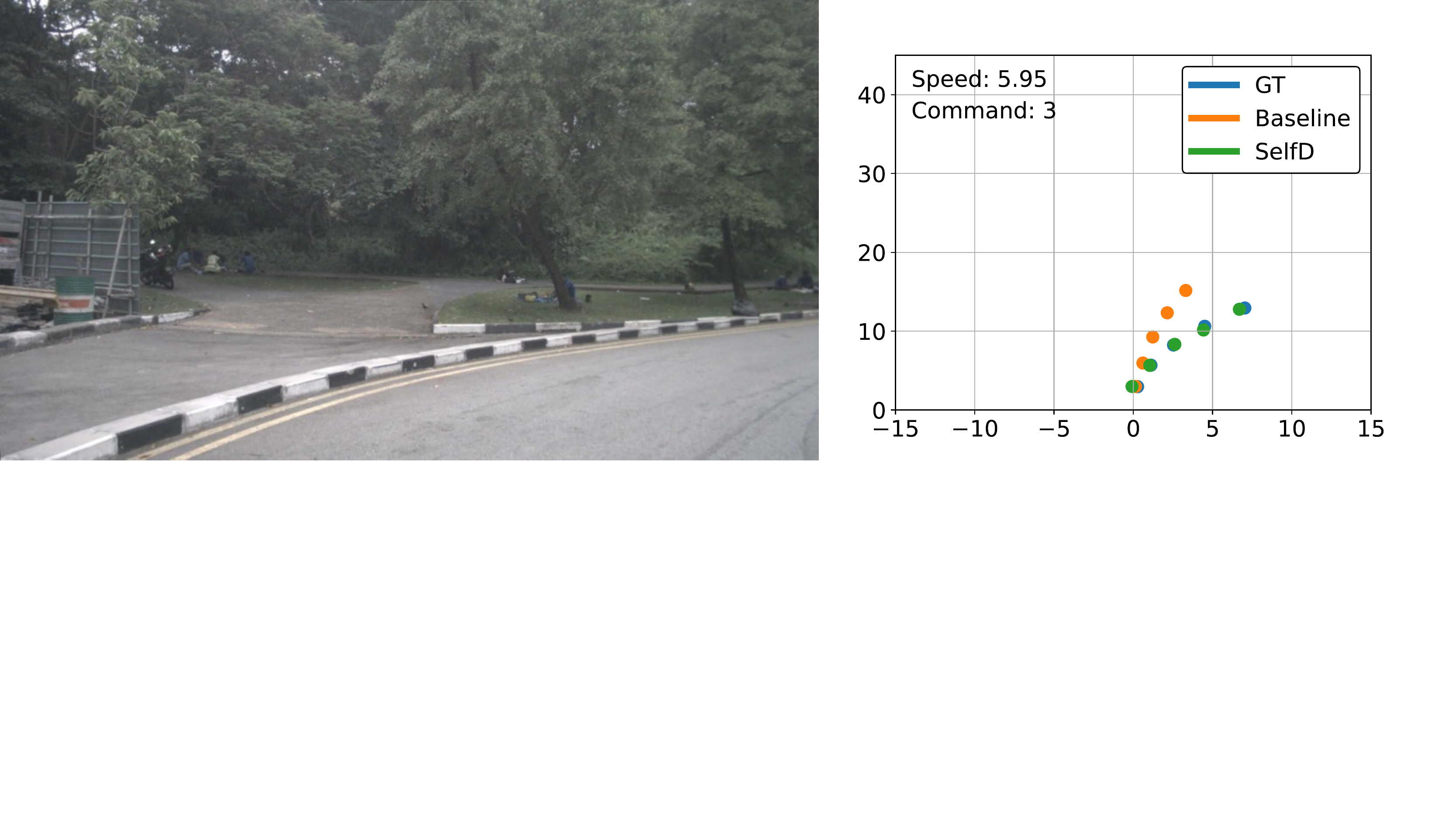} 
       \includegraphics[trim=0cm 8cm 1.5cm 0cm, width=3.3in]{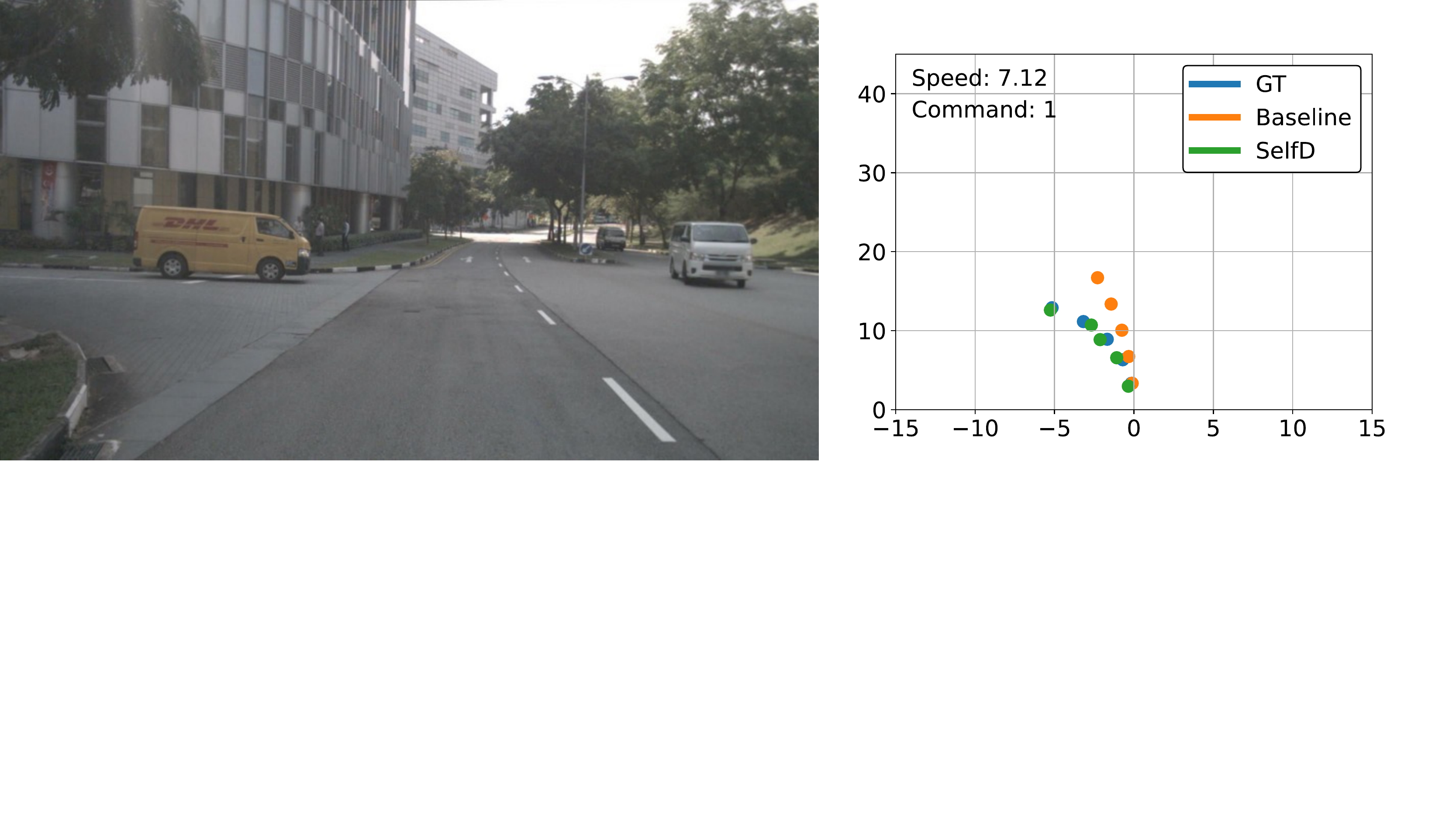}\\
       \includegraphics[trim=0cm 8cm 1.5cm 0cm, width=3.3in]{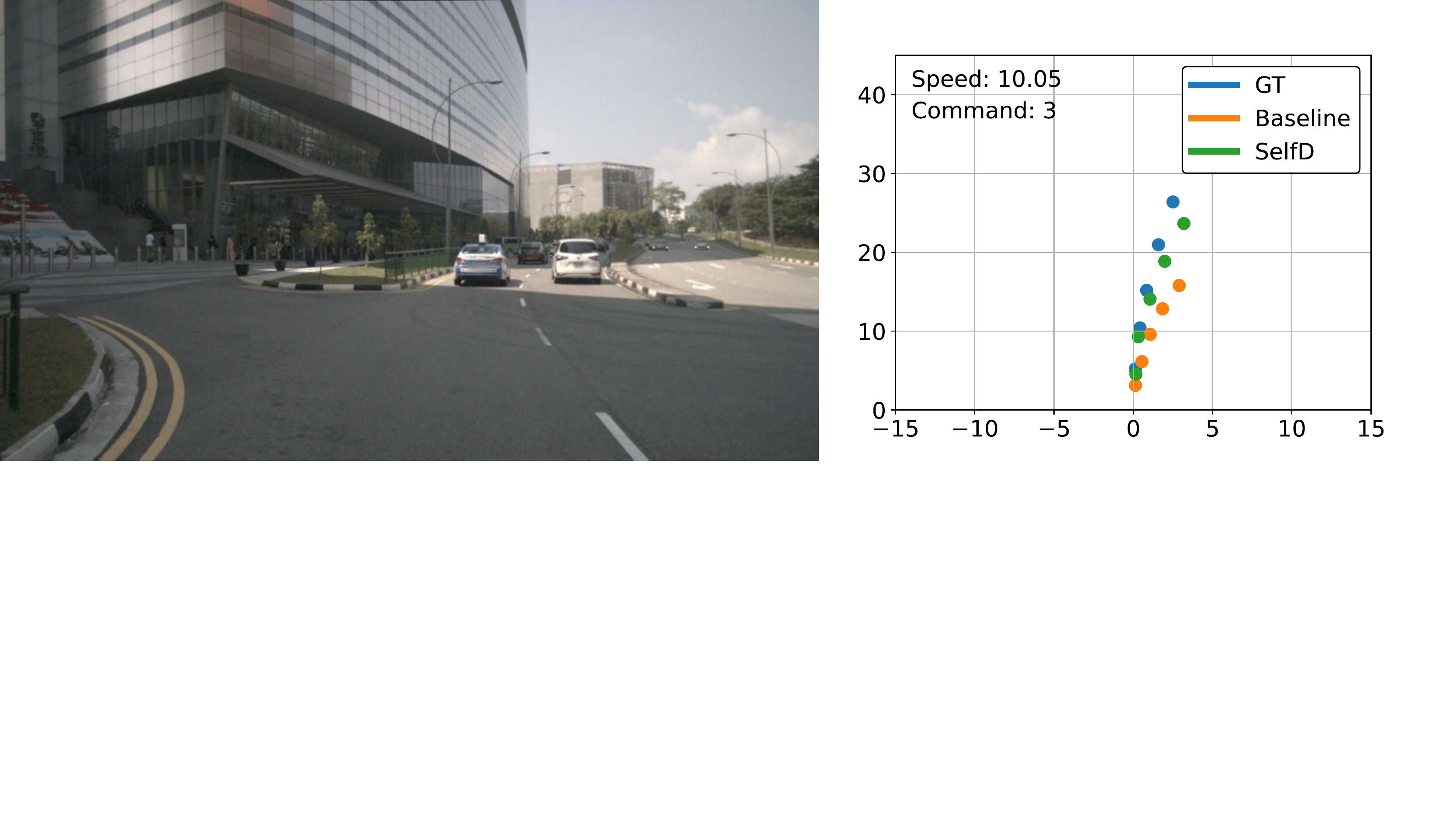} 
       \includegraphics[trim=0cm 8cm 1.5cm 0cm, width=3.3in]{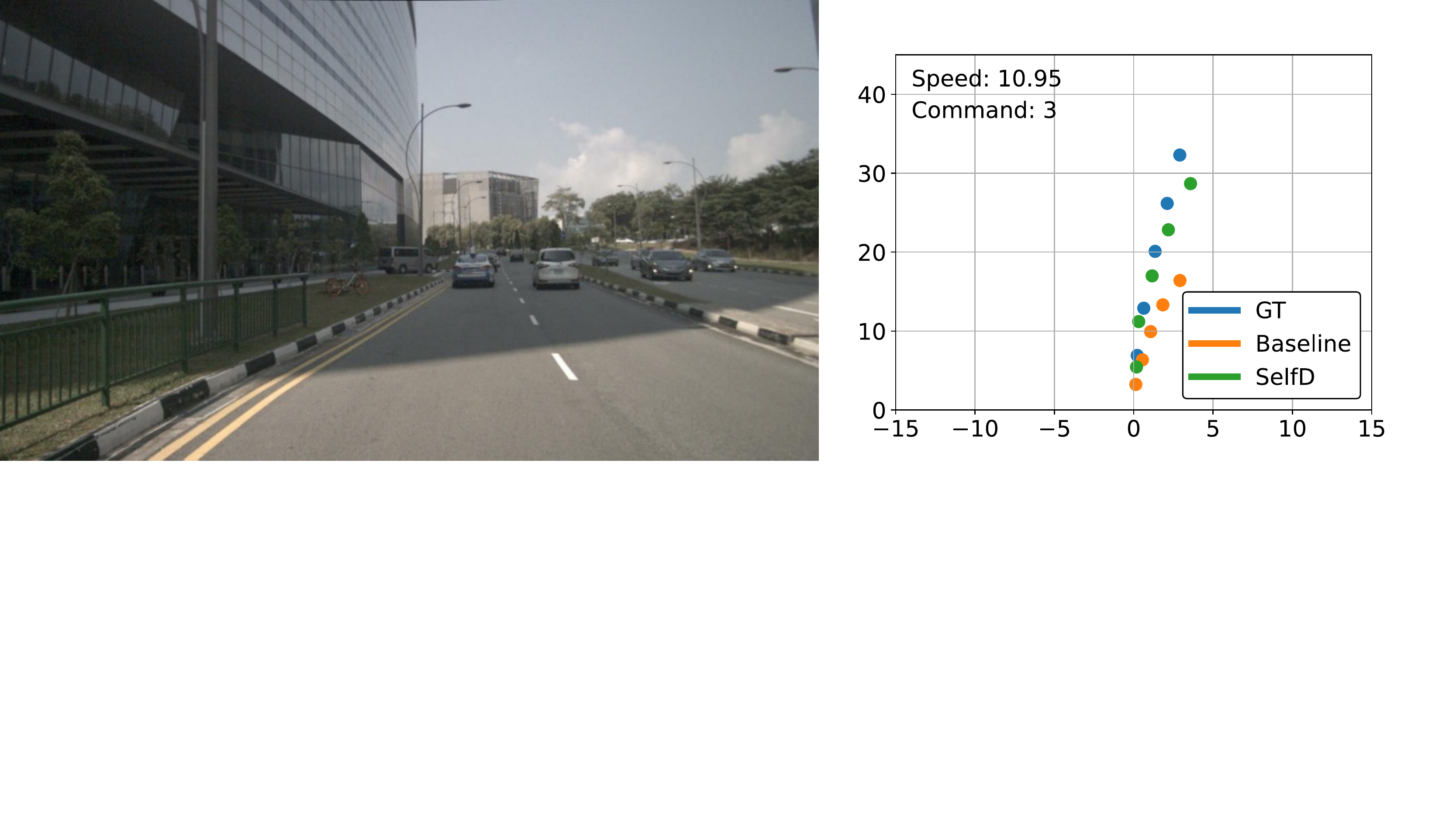}
       
      \caption{\textbf{Qualitative Results.} We plot the performance of SelfD over Waymo (first two rows), Argoverse (middle two rows), and nS-Singapore (bottom two rows). 
      }
   \label{fig:success_qua}
  \end{figure*}

\boldparagraph{Additional Qualitative Results}
We provide additional qualitative results in Fig.~\ref{fig:success_qua} to show a comparison of the proposed SelfD with the baseline model (trained solely on nS-Boston). 
As shown in the figure, the baseline model generally provides as good or worse performance on challenging tasks of navigating across various turn configurations, speeds, maneuvers, and environmental conditions. Interestingly, we find SelfD to robustly handle varying speed ranges, especially at high magnitude. This is depicted in several instance in Fig.~\ref{fig:success_qua}, \eg, example in the top-left corner or the example in the second row and second column. 
\begin{figure*}[!t] 
           \centering
       \includegraphics[trim=0cm 7cm 1.5cm 0cm, width=3.3in]{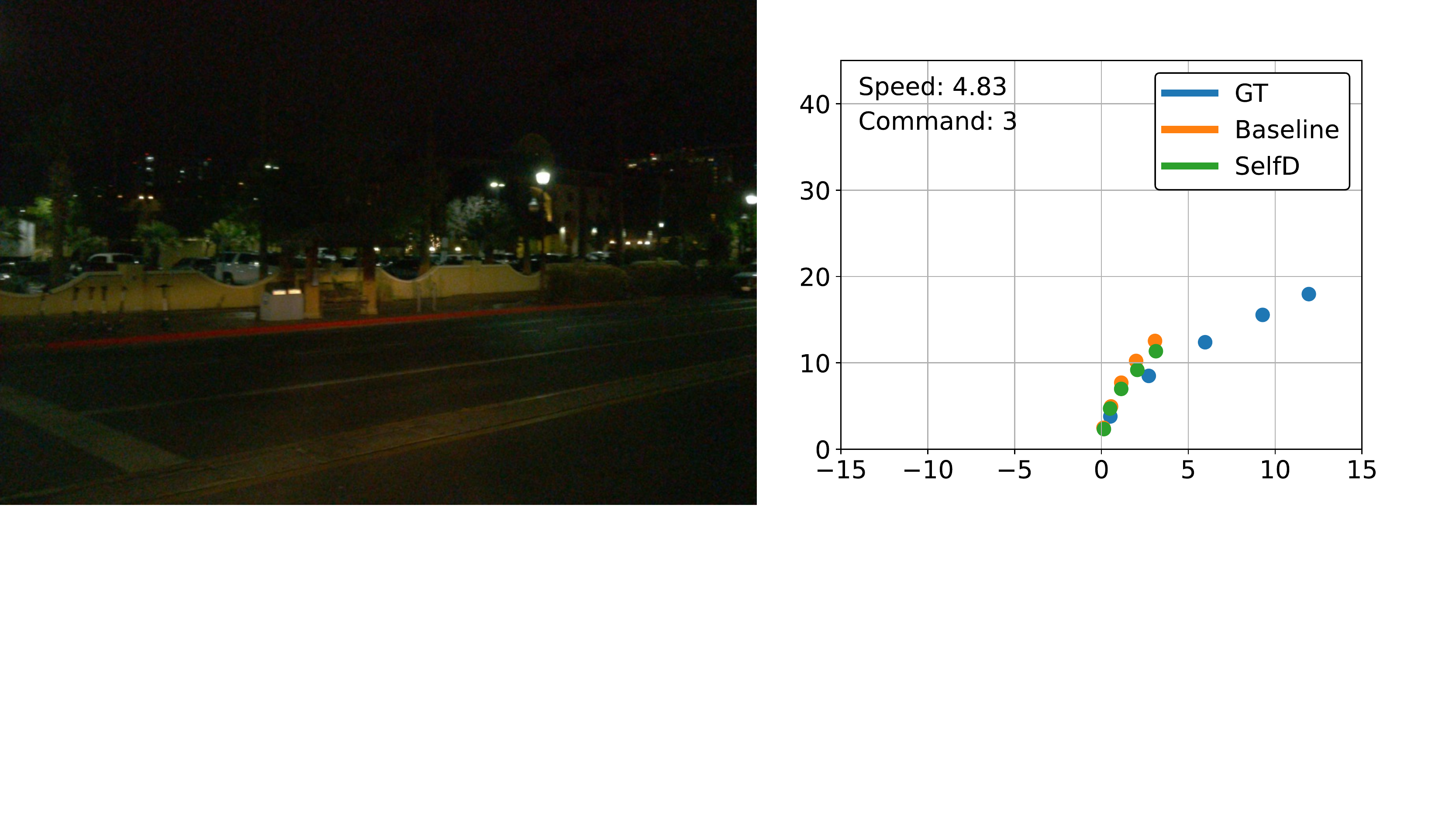} 
          \includegraphics[trim=0cm 7.3cm 1.5cm 0cm, width=3.3in]{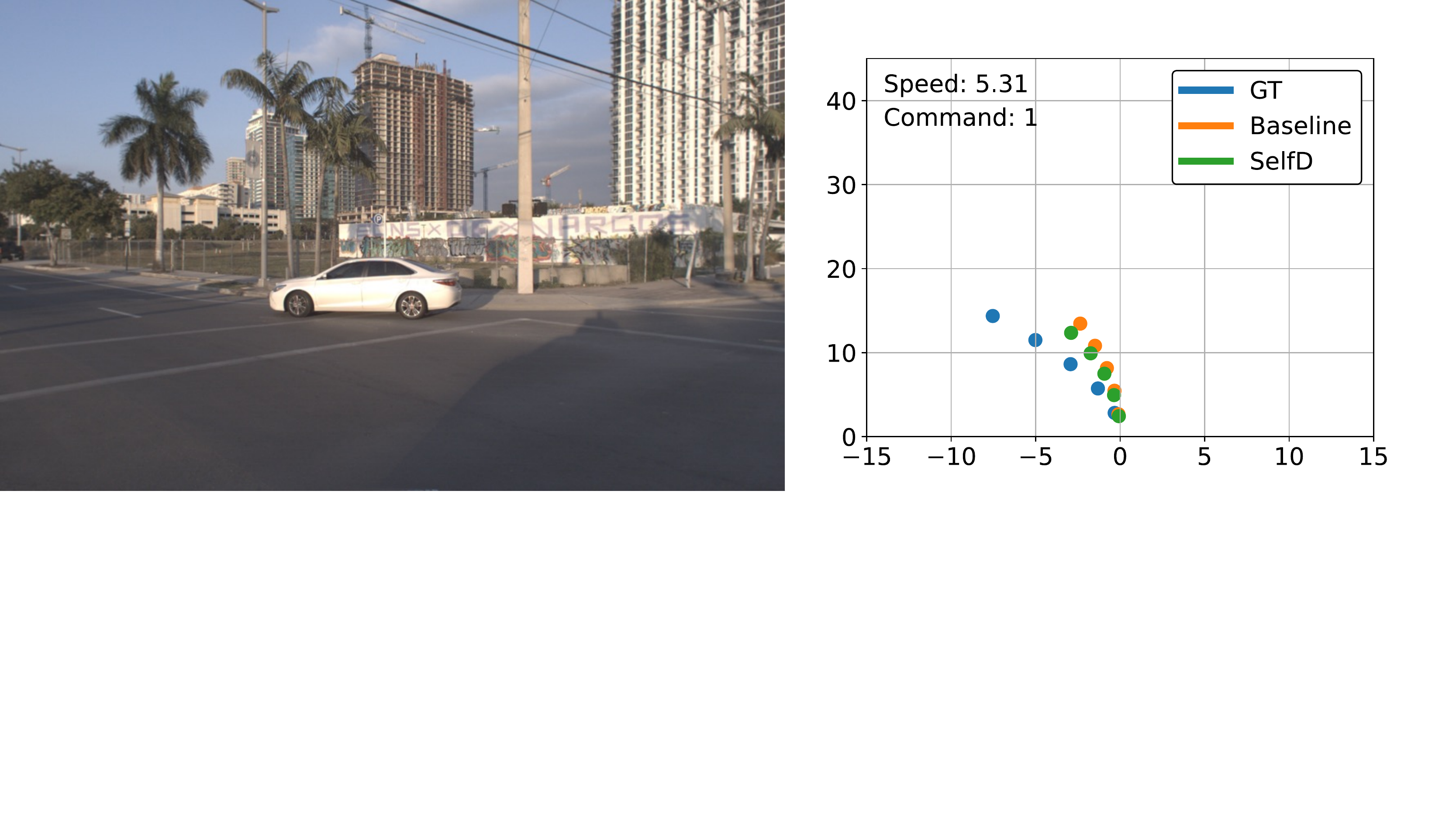}
        \includegraphics[trim=0cm 8cm 1.5cm 0cm, width=3.3in]{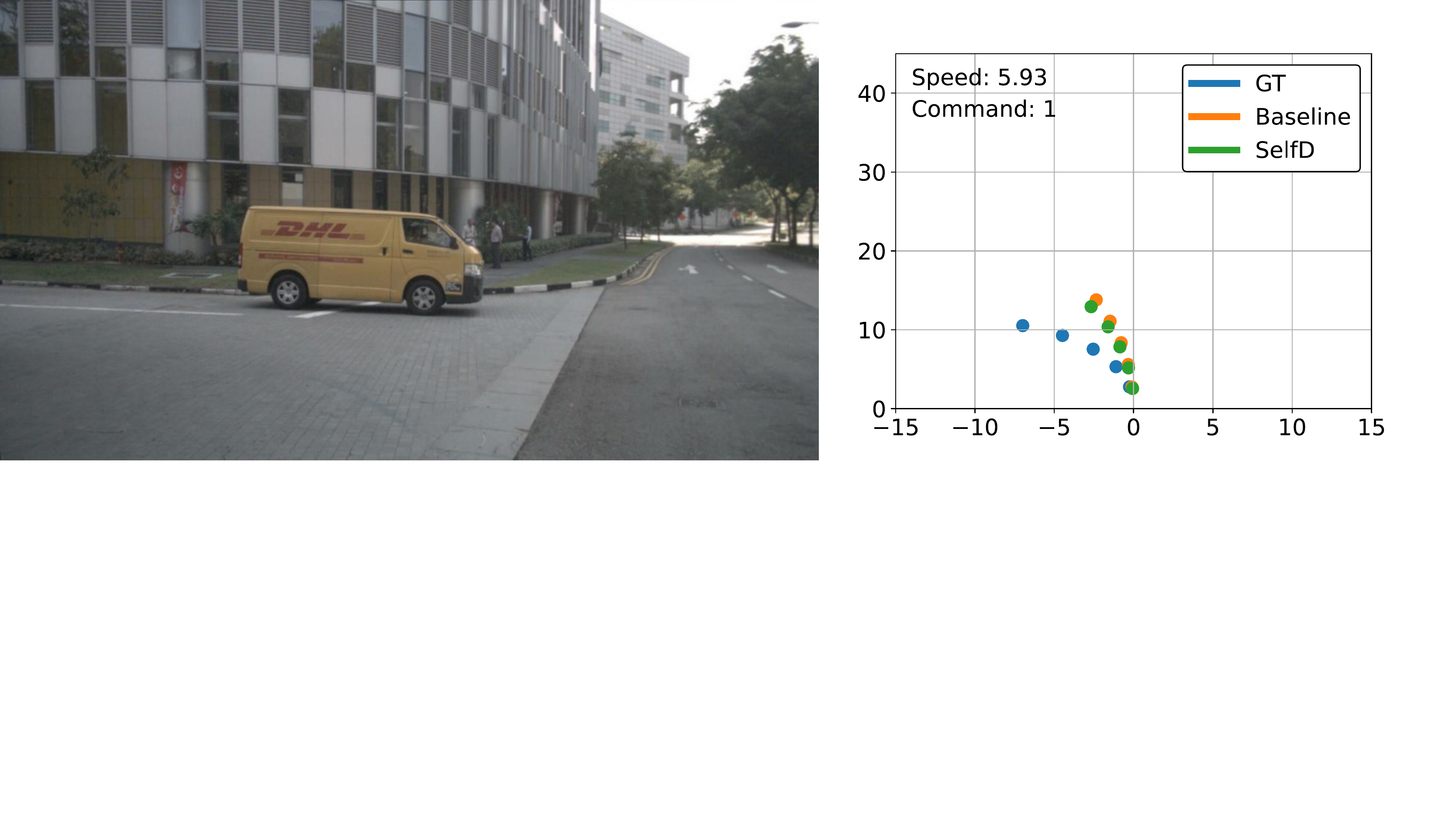}
       \includegraphics[trim=0cm 8cm 1.5cm 0cm, width=3.3in]{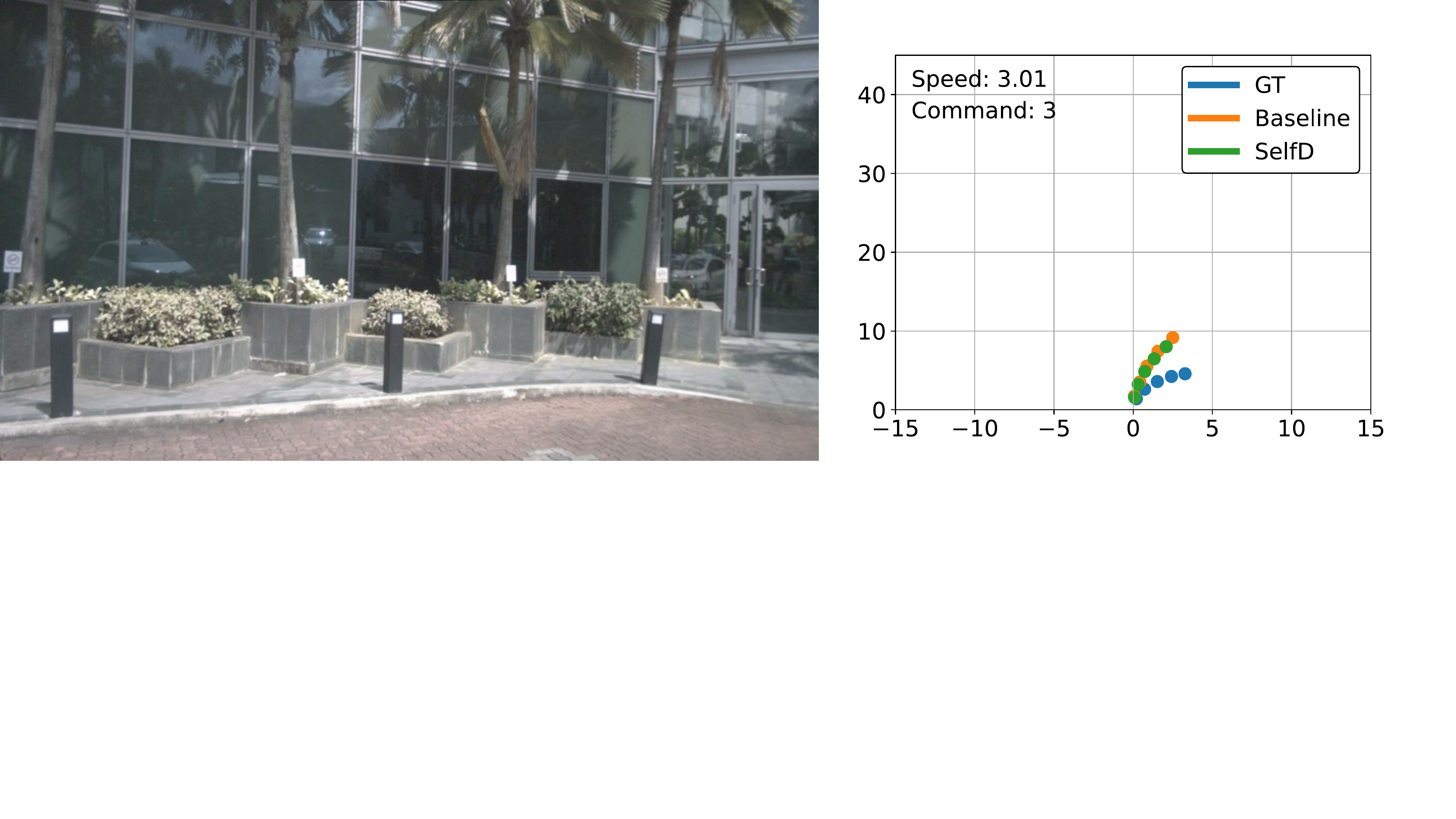}
      \caption{\textbf{Failure Cases.} We visualize BEV waypoint prediction for across datasets (top-left is Waymo, top-right is Argoverse, bottom row is nS-Singapore). While we do observe general ADE and FDE reductions on difficult turn scenarios, the benefits of our current model and YouTube dataset are still limited in such settings, as shown by these difficult examples.  }
         \vspace{-0.4cm}
   \label{fig:failure_qua}
  \end{figure*}

\section*{Appendix C. Limitations}
Finally, we discuss several limitations of our study and model. We begin by considering failure cases as shown in Fig.~\ref{fig:failure_qua}. The visualization specifically focuses on the case of very sharp turns, where we demonstrate the limited utility of the current model. As depicted in the figure, the proposed YouTube-based pre-training pipeline provides very limited improvement in such conditions compared to the baseline. While we followed standard definitions of the conditional commands~\cite{lbc}, it is possible that further improvements in such settings can be made through introducing more fine-grained conditional branches. As adding conditional branches results in a significant increase for the size of the underlying BEV planner model and thus its scalability, we focus on extensively analyzing the three-branch architecture. Nonetheless, while such architecture may learn both sharp and mild turns, we find it to be limited in expressiveness under rare conditions. Fundamentally, the pseudo-labels may also become quite noisy under such instances thus limiting the benefits of the proposed semi-supervised learning step. However, even under such difficult settings, our generalized approach could potentially benefit from additional unlabeled data jointly with an increased model capacity. We leave this to future work. 

A related challenge involves robustness over viewpoints and camera parameters. While prior work generally trains and tests planning models within the same sensor setup, our proposed approach takes a step towards more configuration-invariant navigation. However, the variability in camera configuration among the evaluation datasets used in our study is limited, \ie, compared to completely unconstrained settings. Understanding generalization under such challenges can also benefit from controllable analysis in the future (as was done over weathers and time of day). This can also take our approach closer to demonstrating utility under broader embodied tasks, \ie, beyond on-road driving.

\end{document}